\documentclass{article}

\usepackage{arxiv}

\usepackage[utf8]{inputenc} 
\usepackage[T1]{fontenc}    
\usepackage{url}            
\usepackage{booktabs}       
\usepackage{amsfonts}       
\usepackage{nicefrac}       
\usepackage{microtype}      
\usepackage{lipsum}		
\usepackage{xargs} 
\usepackage{ragged2e}
\usepackage{caption}
\usepackage{xcolor}
\usepackage{colortbl}
\usepackage{xspace}
\usepackage{graphics,amssymb,amsmath,epsfig,color}
\usepackage{graphicx}
\usepackage{multirow}
\usepackage{multicol}
\usepackage{afterpage}
\usepackage{tabularx}
\usepackage{booktabs}
\usepackage{xstring}                     
\usepackage{graphicx}
\usepackage{rotating} 
\usepackage{longtable}
\usepackage{pgfplots}
\usepackage{pgfplotstable}
\usepackage{breakcites}
\usepackage{array}
\usepackage{titlesec}
\newcounter{mytable}\newcounter{myfigure}
\titleformat{\section}
  [hang]
  {\setcounter{mytable}{\value{table}}\setcounter{myfigure}{\value{figure}}
   \setcounter{table}{0}\global\def\thetable{S\arabic{table}}%
   \setcounter{figure}{0}%
   \normalfont\Large\bfseries}
  {\setcounter{table}{\value{mytable}}\setcounter{figure}{\value{myfigure}}
   \global\def\thetable{\arabic{table}}%
   \thesection}
  {2.3ex plus.2ex}
  {}
\usepackage{collcell}
\usepackage[colorinlistoftodos,prependcaption,textsize=tiny]{todonotes}
\usepackage[hidelinks]{hyperref}       

\usepackage[acronym,xindy,nonumberlist]{glossaries}
\makeglossaries

\newcommand*{\phoenix}{RWTH-PHOENIX-Weather\xspace}
\newcommand*{\phoenixs}{PHOENIX\xspace}

\makeatletter
\DeclareRobustCommand\onedot{\futurelet\@let@token\@onedot}
\def\@onedot{\ifx\@let@token.\else.\null\fi\xspace}
\def\eg{{e.g}\onedot} \def\Eg{{E.g}\onedot}
\def\ie{{i.e}\onedot} 
 
\def\etc{{etc}\onedot}

\makeatother

\newcolumntype{C}[1]{>{\centering\let\newline\\\arraybackslash\hspace{0pt}}m{#1}}
\newcolumntype{R}[1]{>{\raggedleft\let\newline\\\arraybackslash\hspace{0pt}}m{#1}}

\title{Quantitative Survey of the State of the Art in\\ Sign Language Recognition}

\author{ \href{https://orcid.org/0000-0002-5555-3900}{\includegraphics[scale=0.06]{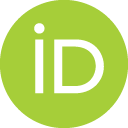}\hspace{1mm}Oscar~Koller}\thanks{\url{https://www.microsoft.com/en-us/research/people/oskoller/}} \\
              Speech and Language\\
              Microsoft\\
              Munich, Germany\\
	\texttt{oscar.koller@microsoft.com} \\
}

\renewcommand{\headeright}{}
\renewcommand{\undertitle}{}

\hypersetup{
pdftitle={Quantitative Survey of the State of the Art in Sign Language
Recognition},
pdfauthor={Oscar Koller},
pdfkeywords={Sign Language Recognition, Survey},
}
\makeglossaries
\newacronym{aam}{AAM}{active appearance model}
\newacronym{am}{AM}{acoustic model}
\newacronym{ann}{ANN}{artificial neural network}
\newacronym{ap}{AP}{average precision}
\newacronym{ar}{ArSL}{Arabic sign language}
\newacronym{argentiniansl}{ArgentSL}{Argentinian sign language}
\newacronym{arsl}{ArSL}{Arabic sign language}
\newacronym{asl}{ASL}{American sign language}
\newacronym{asllrp}{ASLLRP}{American Sign Language Linguistic Research Project}
\newacronym{aslr}{ASLR}{automatic sign language recognition}
\newacronym{asr}{ASR}{automatic speech recognition}
\newacronym{au}{AU}{action unit}
\newacronym{auc}{AUC}{area under curve}
\newacronym{auslan}{Auslan}{Australian sign language}
\newacronym{avsr}{AVSR}{audio-visual speech recognition}
\newacronym{bcc}{BCC}{binary connected components analysis}
\newacronym{bfgs}{BFGS}{Broyden-Fletcher-Goldfarb-Shanno}
\newacronym{bhmm}{BHMM}{Bernoulli \protect\gls{hmm}}
\newacronym{bic}{BIC}{Bayesian information criterion}
\newacronym[\glsshortpluralkey={BLSTMs},\glslongpluralkey={bi-directional long short-term memories},]{blstm}{BLSTM}{bi-directional long short-term memory}
\newacronym{bovw}{BOVW}{bag of visual words}
\newacronym{bow}{BOW}{bag of words}
\newacronym{bsl}{BSL}{British sign language}
\newacronym{cart}{CART}{classification and regression tree}
\newacronym{cbir}{CBIR}{content-based image retrieval}
\newacronym{cc}{CC}{classifier combination}
\newacronym{ce}{CE}{cross-entropy}
\newacronym{cer}{CER}{character error rate}
\newacronym{cmllr}{CMLLR}{constrained maximum-likelihood linear regression}
\newacronym{cnn}{CNN}{convolutional neural network}
\newacronym{csl}{CSL}{Chinese sign language}
\newacronym{cslr}{CSLR}{continuous sign language recognition}
\newacronym{ctc}{CTC}{connectionist temporal classification}
\newacronym{cv}{CV}{computer vision}
\newacronym{czsl}{CzSL}{Czech sign language}
\newacronym{damp}{DAMP}{DARPA Arabic machine-print}
\newacronym{dbn}{DBN}{deep belief network}
\newacronym{dc}{DC}{direct current}
\newacronym{dcs}{DCS}{disparity cost slices}
\newacronym{dct}{DCT}{discrete cosine transform}
\newacronym{dgs}{DGS}{German Sign Language (Deutsche Gebärdensprache)}
\newacronym{dnn}{DNN}{deep neural network}
\newacronym{dog}{DoG}{difference of Gaussians}
\newacronym{dp}{DP}{dynamic programming}
\newacronym{dpf}{GPF}{dynamic partial function}
\newacronym{dpt}{DPT}{dynamic programming tracking}
\newacronym{dsgs}{DSGS}{Swiss German sign language / Deutschschweizerische Gebärdensprache}
\newacronym{dtw}{DTW}{dynamic time warping}
\newacronym{ebw}{EBW}{extended Baum Welch}
\newacronym{eer}{EER}{equal error rate}
\newacronym{emd}{EMD}{earth movers distance}
\newacronym{em}{EM}{expectation maximization}
\newacronym{er}{ER}{error rate}
\newacronym{facs}{FACS}{Facial Action Coding System}
\newacronym{flemishsl}{FlemishSL}{Flemish sign language}
\newacronym{fmllr}{fMLLR}{feature space \protect\glstext{mllr}}
\newacronym{fps}{fps}{frames per second}
\newacronym{fra}{FRA}{frame recognition accuracy}
\newacronym{fsa}{FSA}{finite state automaton}
\newacronym{fsw}{FSW}{Formal and Regular SignWriting}
\newacronym{gdl}{GDL}{glyph dependent length}
\newacronym{ghmm}{GHMM}{Gaussian \protect\gls{hmm}}
\newacronym{gmmhmm}{GMM-HMM}{Gaussian mixture model \protect\gls{hmm}}
\newacronym{gis}{GIS}{generalised iterative scaling}
\newacronym{gmd}{GMD}{Gaussian mixture densities}
\newacronym{gmm}{GMM}{Gaussian mixture model}
\newacronym{gnu}{GNU}{GNU's not Unix}
\newacronym{gpl}{GPL}{General Public License}
\newacronym{gtf}{GTF}{global texture features}
\newacronym{gsl}{GSL}{Greek sign language}
\newacronym{hamnosys}{HamNoSys}{Hamburg Notation System}
\newacronym{hdm}{HDM}{histogram distortion model}
\newacronym{hksl}{HKSL}{Hong Kong sign language}
\newacronym{hltpr}{HLTPR}{Human Language Technology and Pattern Recognition}
\newacronym{hmm}{HMM}{hidden Markov model}
\newacronym{hog}{HoG}{histogram of oriented gradients}
\newacronym{hp}{HP}{hand position}
\newacronym{hsv}{HSV}{hue-saturation-value}
\newacronym{ht}{HT}{hand trajectory}
\newacronym{htk}{HTK}{Hidden Markov Model Toolkit}
\newacronym{hv}{HV}{hand velocity}
\newacronym{hwr}{HWR}{handwriting recognition}
\newacronym{iapr}{IAPR}{International Association for Pattern  Recognition}
\newacronym{icdar}{ICDAR}{International Conference on Document Analysis and Recognition}
\newacronym{icfhr}{ICFHR}{International Conference on Frontiers in Handwriting Recognition}
\newacronym{icr}{ICR}{intelligent character recognition}
\newacronym{idf}{IDF}{inverse document frequency}
\newacronym{idm}{IDM}{image distortion model}
\newacronym{ifh}{IFH}{invariant feature histogram}
\newacronym{ilsvrc}{ILSVRC}{ImageNet Large-Scale Visual Recognition Challenge}
\newacronym{imerr}{IER}{image error rate}
\newacronym{indiansl}{IndianSL}{Indian sign language}
\newacronym{indonesiansl}{IndoSL}{Indonesian sign language}
\newacronym{ip}{IP}{interest point}
\newacronym{irma}{IRMA}{Image Retrieval in Medical Applications}
\newacronym{irishsl}{IrishSL}{Irish sign language}
\newacronym{is}{IS}{International Sign}
\newacronym{isl}{ISL}{Irish Sign Language}
\newacronym{iwr}{IWR}{intelligent word recognition}
\newacronym{iwslt}{IWSLT}{international workshop on spoken language  translation}
\newacronym{jsd}{JSD}{Jensen-Shannon divergence}
\newacronym{jsl}{JSL}{Japanese sign language}
\newacronym{kl}{KL}{Kullback-Leibler}
\newacronym{klt}{KLT}{Kanade-Lucas-Tomasi}
\newacronym{kurdishsl}{KurdishSL}{Kurdish sign language}
\newacronym{ksl}{KSL}{Korean sign language}
\newacronym{knn}{KNN}{$k$ nearest neighbor}
\newacronym{krsl}{K-RSL}{Kazakh-Russian sign language}

\newacronym{l1o}{L1O}{leaving one out}
\newacronym{lbfgs}{LBFGS}{limited memory \protect\gls{bfgs}}
\newacronym{lbg}{LBG}{Linde-Buzo-Gray}
\newacronym{lbp}{LBP}{local binary pattern}
\newacronym{lbw}{LBW}{Lancaster-Oslo-Bergen, Brown and Wellington}
\newacronym{lda}{LDA}{linear discriminant analysis}
\newacronym{ldc}{LDC}{Linguistic Data Consortium}
\newacronym{lf}{LF}{local feature}
\newacronym{libras}{Libras}{Brazilian sign language / Lingua Brasileira de sinais}
\newacronym{lis}{LIS}{Italian sign language / Lingua Italiana dei segni}
\newacronym{lse}{LSE}{Spanish sign language / Lengua de signos española}
\newacronym{lm}{LM}{language model}
\newacronym{lob}{LOB}{Lancaster-Oslo-Bergen}
\newacronym{lsf}{LSF}{Langue des Signes Fran{\c}aise}
\newacronym[\glsshortpluralkey={LSTMs},\glslongpluralkey={long short-term memories},]{lstm}{LSTM}{long short-term memory}
\newacronym{lvcsr}{LVCSR}{large vocabulary continuous speech recognition}
\newacronym{map}{MAP}{mean average  precision}
\newacronym{malaysl}{MalaySL}{Malaysian sign language}
\newacronym{mce}{MCE}{minimum classification error}
\newacronym{mds}{MDS}{multi-dimensional scaling}
\newacronym{me}{ME}{maximum entropy}
\newacronym{mexicansl}{MexicanSL}{Mexican sign language}
\newacronym{mfdi}{MFDI}{mean face difference image}
\newacronym{mil}{MIL}{multiple instance learning}
\newacronym{mir}{MIR}{Mallinkrodt Institute of Radiology}
\newacronym{mit}{MIT}{Massachusetts Institute of Technology}
\newacronym{mle}{MLE}{model length estimation}
\newacronym{mllr}{MLLR}{maximum likelihood linear regression}
\newacronym{ml}{ML}{maximum likelihood}
\newacronym{mlpghmm}{MLP-GHMM}{\protect\gls{mlp}-\protect\gls{ghmm}}
\newacronym{mlp}{MLP}{multi-layer perceptron}
\newacronym{mmac}{MMAC}{Multi-Modal Arabic Corpus}
\newacronym{mmiconf}{MMI-conf}{confidence-based \protect\gls{mmi}}
\newacronym{mmi}{MMI}{maximum mutual information}
\newacronym{mmmiconf}{M-MMI-conf}{confidence-based \protect\gls{mmmi}}
\newacronym{mmmi}{M-MMI}{margin-based \protect\gls{mmi}}
\newacronym{mmpeconf}{M-MPE-conf}{confidence-based \protect\gls{mmpe}}
\newacronym{mmpe}{M-MPE}{margin-based \protect\gls{mpe}}
\newacronym{mnist}{MNIST}{modified \protect\gls{nist}}
\newacronym{mpeconf}{MPE-conf}{confidence-based \protect\gls{mpe}}
\newacronym{mpeg}{MPEG}{Moving Picture Experts Group}
\newacronym{mpe}{MPE}{minimum phone error}
\newacronym{mser}{MSER}{maximally stable extremal regions}
\newacronym{msrc}{MSRC}{Microsoft Research Cambridge}
\newacronym{mt}{MT}{machine translation}
\newacronym{mwe}{MWE}{minimum word error}
\newacronym{ngt}{NGT}{Dutch sign language / Nederlandse Gebaren Taal}
\newacronym{nist}{NIST}{National Institute for Standards and Technology}
\newacronym{nn}{NN}{nearest neighbour}
\newacronym{numa}{NUMA}{non-uniform memory architecture}
\newacronym{oao}{OAO}{one against one}
\newacronym{oatr}{OATR}{one against the rest}
\newacronym{ocr}{OCR}{optical character recognition}
\newacronym{odel}{DEL}{object deletion}
\newacronym{oerr}{OER}{object error rate}
\newacronym{ohsu}{OHSU}{Oregon Health and Science University}
\newacronym{oins}{INS}{object insertion}
\newacronym{oov}{OOV}{out of vocabulary}
\newacronym{osub}{SUB}{object substitution}
\newacronym{p2dhmm}{P2DHMM}{pseudo two-dimensional \protect\gls{hmm}}
\newacronym{pascal}{PASCAL}{Pattern Analysis, Statistical Modelling and Computational Learning}
\newacronym{patdb}{PATDB}{Printed Arabic Text Database}
\newacronym{paw}{PAW}{piece of Arabic word}
\newacronym{pca}{PCA}{principal components analysis}
\newacronym{pc}{PC}{personal computer}
\newacronym{pdf}{PDF}{probablity density function}
\newacronym{pdm}{PDM}{point density model}
\newacronym{peir}{PEIR}{Pathology Education Instructional Resources}
\newacronym{persiansl}{PersianSL}{Persian sign language}
\newacronym{poicaam}{POICAAM}{project-out inverse-compositional \protect\gls{aam}}
\newacronym{pp}{PP}{perplexity}
\newacronym{psl}{PolishSL}{Polish sign language}
\newacronym{qbve}{QBVE}{query by visual example}
\newacronym{qs}{QS}{quotient of sums}
\newacronym{rampn}{RAMP-N}{\protect\Gls{rwth} Arabic Machine-Print Newspaper}
\newacronym{rasr}{RASR}{RWTH Aachen University Speech Recognizer}
\newacronym{rast}{RAST}{recognition by adaptive subdivision of the transformation space}
\newacronym{rbf}{RBF}{radial basis function}
\newacronym{relu}{ReLU}{rectified linear unit}
\newacronym{rgb}{RGB}{red green blue}
\newacronym{rimes}{RIMES}{Reconnaissance et Indexiation de donn\'ees Manuscrites et de fac simil\'ES}
\newacronym{rnn}{RNN}{recurrent neural network}
\newacronym{roc}{ROC}{receiver operating characteristic}
\newacronym{roi}{ROI}{region-of-interest}
\newacronym{rprop}{RProp}{resilient backpropagation}
\newacronym{rs}{RS}{relevance score}
\newacronym{russiansl}{RussianSL}{Russian sign language}
\newacronym{rsv}{RSV}{retrieval status value}
\newacronym{rwth-asr}{RWTH ASR}{RWTH Aachen University Speech Recognition}
\newacronym{rwth-ocr}{RWTH OCR}{RWTH Aachen University Optical Character Recognition}
\newacronym{rwth}{RWTH}{RWTH Aachen University}
\newacronym{sat}{SAT}{speaker adaptive training}
\newacronym{see}{SEE}{signed exact English}
\newacronym{sgd}{SGD}{stochastic gradient descent}
\newacronym{sift}{SIFT}{scale invariant feature transformation}
\newacronym{signspeak}{SignSpeak}{SignSpeak}
\newacronym{slt}{SLT}{sign language translation}
\newacronym{slr}{SLR}{sign language recognition}
\newacronym{sl}{SL}{sign language}
\newacronym{smt}{SMT}{statistical machine translation}
\newacronym{ssd}{SSD}{sum of squared distances}
\newacronym{ssl}{SSL}{Swedish sign language}
\newacronym{surf}{SURF}{speeded-up robust features}
\newacronym{svd}{SVD}{singular value decomposition}
\newacronym{svm}{SVM}{support vector machine}
\newacronym{sv}{SV}{support vector}
\newacronym{swv}{SWV}{supervised writing variants}
\newacronym{tcstar}{TC-STAR}{Technology and Corpora for Speech to  Speech Translation}
\newacronym{tc}{TC}{technical committee}
\newacronym[\glslongpluralkey={time distortion penalties},]{tdp}{TDP}{time distortion penalty}
\newacronym{tamilsl}{TamisSL}{Tamil sign language}
\newacronym{ter}{TER}{tracking error rate}
\newacronym{tet}{TET}{PDFlib Text Extraction Toolkit}
\newacronym{tfidf}{TF/IDF}{text frequency/inverse document frequency}
\newacronym{tf}{TF}{text frequency}
\newacronym{trap}{TRAP}{TempoRAl Pattern}
\newacronym{trec}{TReC}{Text Retrieval Conference}
\newacronym{tsl}{TSL}{Turkish sign language}
\newacronym{ttr}{TTR}{types/token ratio}
\newacronym{twsl}{TaiwanSL}{Taiwanese sign language}
\newacronym{uibk}{UIBK}{University of Innsbruck}
\newacronym{upv}{UPV}{Universidad Polit{\'e}cnica de Valencia}
\newacronym{url}{URL}{unique resource locator}
\newacronym{usps}{USPS}{US Postal Service}
\newacronym{uw}{UW}{University of Washington}
\newacronym{vjd}{VJD}{Viola \& Jones Detector}
\newacronym{vjt}{VJT}{Viola \& Jones Tracker}
\newacronym{vj}{VJ}{Viola \& Jones}
\newacronym{vm}{VM}{visual model}
\newacronym{voc}{VOC}{visual object classes challenge}
\newacronym{vsa}{VSA}{visual speaker adaptation}
\newacronym{vts}{VTS}{virtual training sample}
\newacronym{wat}{WAT}{writer adaptive training}
\newacronym{wer}{WER}{word error rate}
\newacronym{wta}{WTA}{winner-takes-all}
\newacronym{zow}{ZOW}{zero-order warping}

\newglossaryentry{continuous}{
  name={continuous},%
  description={Specifies the nature of sign language data sets that
    encompass long phrases or full sentences as opposed to single, isolated signs}}

\newglossaryentry{gloss}{%
  name={gloss},%
  description={Expressing the meaning of signs with words of
    a spoken language}}

\newglossaryentry{intrusive}{
  name={intrusive},%
  description={Specifies the capturing of sign language data sets that
    requires the signer to wear specific measuring devices such as
    gloves or trackers}}

\newglossaryentry{isolated}{
  name={isolated},%
  description={Specifies the nature of sign language data sets that
    only encompass single signs as opposed to long phrases or full sentences}}

\newglossaryentry{non-intrusive}{
  name={non-intrusive},%
  description={Specifies the capturing of sign language data sets that
    does not require the signer to wear specific measuring devices such as
    gloves or trackers}}

\newglossaryentry{parameter}{
  name={parameter},%
  description={Each sign consists of a set of parameters. We
    distinguish manual and non-manual parameters. Hand shape,
    orientation, location and movement are the four
    manual parameter, while non-manual parameters include head and
    body posture, facial expression, eye gaze and mouth patterns}}

\newglossaryentry{vocabulary}{%
  plural={vocabularies},
  name={vocabulary},%
  description={The set of unique signs (or words) that occur in a
    dataset. Typically, statistical recognition systems are limited to recognize a
    specific set of words: the vocabulary}
}



\begin{document}
\maketitle

\begin{abstract}
  This work presents a meta study covering around 300 published sign
  language recognition papers with over 400 experimental results. It 
  includes most papers between the start of the field in 1983 and 2020. Additionally, it
  covers a fine-grained analysis on over 25 studies that have
  compared their recognition approaches on \phoenix 2014, the standard benchmark task of the field. 
  Research in the domain of sign language recognition has progressed
  significantly in the last decade, reaching a point where the task
  attracts much more attention than ever before. This study compiles the state
  of the art in a concise way to help advance the field and reveal
  open questions. Moreover, all of this meta study's source data is made public, easing future work with it and further expansion.
  The analyzed papers have been manually labeled with a set of categories.
  The data reveals many insights, such as, among others, shifts in the field from intrusive
  to non-intrusive capturing, from local to global features and the
  lack of non-manual parameters included in medium and larger
  vocabulary recognition systems. Surprisingly, \phoenix with a
  vocabulary of 1080 signs represents the only resource for large
  vocabulary continuous sign language recognition benchmarking world wide. 
  
\end{abstract}


\section{Introduction}
Since recently, automatic sign language recognition experiences significantly more attention by the community.
The number of published studies, but also the quantity of available data sets is
increasing. 
This work aims at providing an overview of the field following a
quantitative meta-study approach. For that, the author covered the most relevant 300 published
studies, since the earliest known work~\cite{grimes_digital_1983}. The
300 analyzed recognition studies
have been manually labeled based on their basic recognition
characteristics such as modeled \gls{vocabulary} size, the
number of contributing signers, the tackled sign language and additional details, such as the quality of the
employed data set (\eg if it covers \gls{isolated} or \gls{continuous} sign
language), the available input data type (\eg if provides colors as
well as depth information or specific measuring devices for tracking
body parts) and the employed sign language modalities and features
(\eg which of the manual and non-manual sign language parameters have been
explicitly modeled and which additional features are employed).
Based on this data, extensive analysis is presented by creating 
 graphics and tables that relate specific
characteristics, visualize correlations, highlight
short-comings and allow to create proven hypotheses.
Beyond that, this
work focuses on the \phoenix data set, which has evolved to currently be the standard
benchmark data set of the sign language recognition field. We provide
a detailed structured view
comparing over 25 research studies that have evaluated their approaches on
the \phoenix corpus. We track the employed neural
architectures, the training style, the employed losses and the data
augmentation of all covered studies and present it in a unified table
jointly with the achieved performance.
The raw data of this work is made publicly
available\footnote{\url{https://github.com/oskoller/sign-language-state-of-the-art}}.
As such, this paper makes the following contributions:
\begin{itemize}
\item Extensive quantitative structured data covering a large part of
  the sign language recognition research is made publicly available.
\item First sign language recognition meta study, providing quantitative insights and analysis of the state of the art.
\item First overview and in-depth analysis of all published papers
  that have compared their proposed recognition systems on \phoenixs 2014, the standard benchmark of the field.
\end{itemize}

In the following, we will start 
 in
Section~\ref{sec:state-of-the-art} to dive into the
analysis and present the general development of the field, followed by looking into the
available input data used for modeling in
Section~\ref{sec:type-input-data} and the chosen sign language modalities
and features to be modeled in
Section~\ref{sec:modeled-sign-language-modalities}.
In Section~\ref{sec:before-and-after-2015}, we point out the
differences of the research landscape before and after 2015. We
compare the studies and investigate general sign language recognition
trends as manifested on the \phoenix 2014 benchmark data set in
Section~\ref{sec:phoenix-2014-analysis}.
Finally, we conclude this paper with Section~\ref{sec:conclusion}. The
full data table can be found in the appendix.





%
%
%
\section{Analysis of the State of the Art}
\label{sec:state-of-the-art}
%
%
%

%
%
%
%
Figure~\ref{fig:number_of_results_by_year} shows the number of
published \gls{isolated} and \gls{continuous} recognition results in blocks of
five years up until 2020. We see that growth looks exponential for  \gls{isolated}
studies, while being close to linear for \gls{continuous} studies. This
may reflect the difficulty of the \gls{continuous} recognition
scenario and also the scarcity of available training corpora. On average it seems that
there are at least twice as many studies published using \gls{isolated}
sign language data.

However, 
Figure~\ref{fig:fig:number_of_results_by_vocabulary}, which shows the
number of \gls{isolated} and \gls{continuous} recognition results aggregated by
\gls{vocabulary} size, reveals that the vast majority
of the \gls{isolated} sign language recognition works model a very
limited amount of signs only (\ie. below 50 signs). This is not the
case when comparing \gls{continuous} sign language recognition, where the
overall studies more or less evenly spread across all sign
\glspl{vocabulary} (with exception of 500-1000 signs due to lack of
available corpora).

Table~\ref{tab:vocab_by_year} provides a more detailed perspective on
the same data: Here, the number of published results is shown per year
and per \gls{vocabulary} range. In the middle and lower parts of the table,
we see this information for \gls{isolated} and \gls{continuous} results,
respectively, while in the top part of the table it is provided
jointly for both data qualities. As in
Figure~\ref{fig:number_of_results_by_year} and
\ref{fig:fig:number_of_results_by_vocabulary}, we note that overall the number of
studies increases over the years. However, we also see that this trend
is true for the smallest and medium \gls{vocabulary} (below 50 signs and
between 200 and 1000 signs) only. The large \gls{vocabulary} tasks (over
1000 signs) have been low until year 2015 and following. When looking
at the \gls{continuous} studies only (lower part of
Table~\ref{tab:vocab_by_year}), we see that large \gls{vocabulary} ($>1000$ signs) and
50-200 \gls{vocabulary} tasks have experienced a large gain in the number of
published results since 2015. This can be explained with the community
focusing on two benchmark corpora since then
(\cite{koller_continuous_2015} with a \gls{vocabulary} of 1080 signs
and \cite{huang_videobased_2018} with a \gls{vocabulary} of 178).  
 
\newcommand\histwidth{7.5cm}
\begin{figure}[tbp]
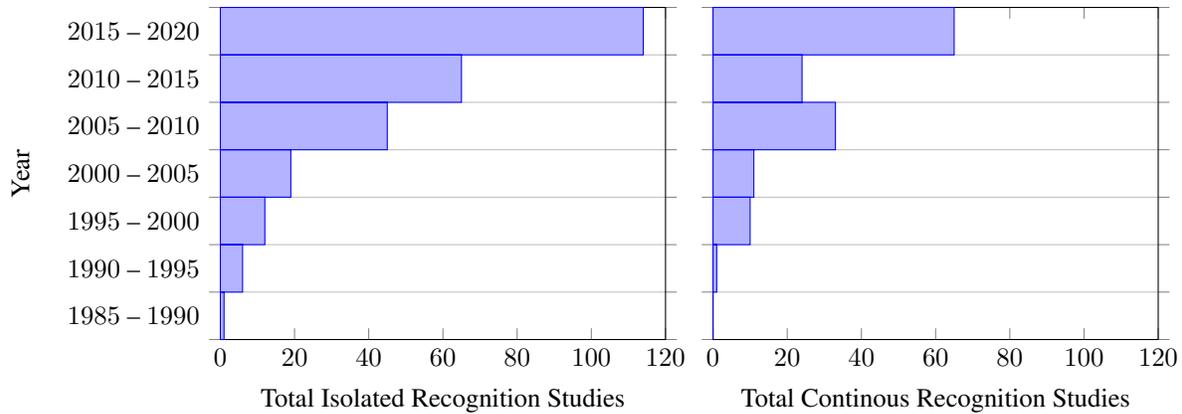

  \centering
  \setlength{\tabcolsep}{-2pt}
  \renewcommand{\arraystretch}{-1} 
  \begin{tabular}{cc}
    \begin{tikzpicture}
      \begin{axis}[
        width=\histwidth,
        height=6cm,
        xmin=0,xmax=120,
        xlabel=Total Isolated Recognition Studies,
        enlargelimits=false,
        ytick=data,
        yticklabel interval boundaries,
        y label style={at={(axis description cs:-0.2,0.5)},anchor=south},
        xbar interval,
        ylabel=Year,
        /pgf/number format/.cd,
        use comma,
        1000 sep={},
        ]
        \addplot
        coordinates
        {\input{histogram-year-iso.cnt} };
      \end{axis}
    \end{tikzpicture} &
    \begin{tikzpicture}
      \begin{axis}[
        width=\histwidth,
        height=6cm,
        xmin=0,xmax=120,
        xlabel=Total Continous Recognition Studies,
        enlargelimits=false,
        xbar interval,
        yticklabels={,,}
        ]
        \addplot
        coordinates
        {\input{histogram-year-cont.cnt} };
      \end{axis}
    \end{tikzpicture} \\
  \end{tabular}
  \caption{Showing the number of published recognition results between 1983 and
    2020.}
  \label{fig:number_of_results_by_year}
\end{figure}

\begin{figure}[tbp]
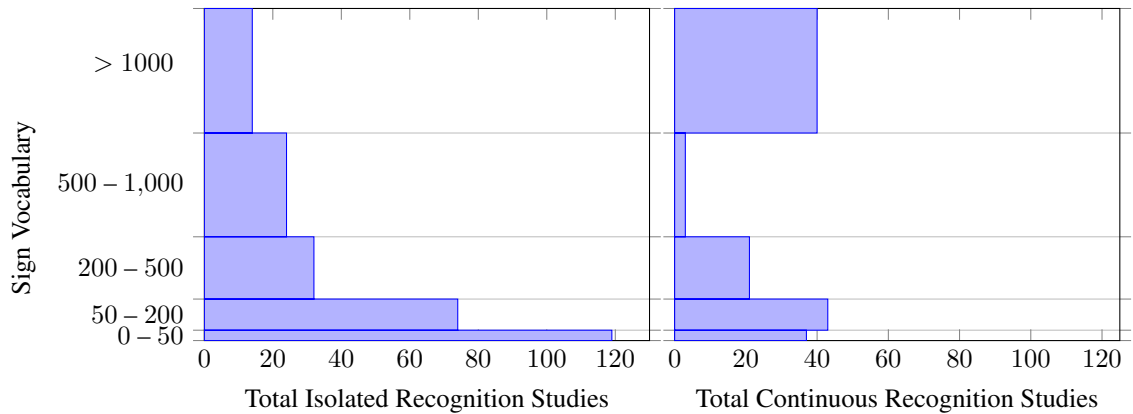

  \centering
  \setlength{\tabcolsep}{-3pt}
  \renewcommand{\arraystretch}{-1} 
  \begin{tabular}{cc}
    \pgfplotsset{
    overwrite last y tick label/.style={
    every y tick label/.append style={alias=lasttick},
    extra description/.append code={
    \fill [white] (lasttick.north west) ++(0pt,-\pgflinewidth) rectangle (lasttick.south east);         
    \node [anchor=base] at (lasttick) {#1};}
    },
    overwrite last y tick label/.default={~ ~ ~ ~$>1000$}
    }
    \begin{tikzpicture}
      \begin{axis}[
        width=\histwidth,
        height=6cm,
        xmin=0,xmax=130,
        ymin=0,
        overwrite last y tick label,
        xlabel=Total Isolated Recognition Studies,
        enlargelimits=false,
        ytick=data,
        yticklabel interval boundaries,
        y label style={at={(axis description cs:-0.15,0.4)},anchor=south},
        xbar interval,
        ylabel=Sign Vocabulary,
        ]
        \addplot
        coordinates
        {\input{histogram-signvocab-iso.cnt} };
      \end{axis}
    \end{tikzpicture}&%
    \begin{tikzpicture}
      \begin{axis}[
        width=\histwidth,
        height=6cm,
        xmin=0, xmax=125,
        xlabel=Total Continuous Recognition Studies,
        enlargelimits=false,
        xbar interval, yticklabels={,,}
        ]
        \addplot
        coordinates
        {\input{histogram-signvocab-cont.cnt} };
      \end{axis}
    \end{tikzpicture} \\
  \end{tabular}
  \caption{Showing the number of published results between 1983 and
    2020 and the size of their modeled sign vocabulary.}
  \label{fig:fig:number_of_results_by_vocabulary}
\end{figure}

\pgfplotstableset{
  string type,
  col sep=&,
  /color cells/min/.initial=0,
  /color cells/max/.initial=1000,
  /color cells/textcolor/.initial=,
  %
  color cells/.code={%
    \pgfqkeys{/color cells}{#1}%
    \pgfkeysalso{%
      postproc cell content/.code={%
        %
        \message{2ndchecking \pgfplotstablecol \pgfplotstablecolname}%
        \begingroup%
        %
        %
        \pgfkeysgetvalue{/pgfplots/table/@preprocessed cell content}\value%
        \pgfmathfloatparsenumber{\value}%
        \pgfmathfloattofixed{\pgfmathresult}%
        \let\value=\pgfmathresult%
        %
        \pgfplotscolormapaccess%
        [\pgfkeysvalueof{/color cells/min}:\pgfkeysvalueof{/color cells/max}]%
        {\value}%
        {\pgfkeysvalueof{/pgfplots/colormap name}}%
        %
        \pgfkeysgetvalue{/pgfplots/table/@cell content}\typesetvalue%
        \message{row \pgfplotstablerow col \pgfplotstablecol \space matches #1...^^J \pgfplotstablecolname}%
        \pgfkeysgetvalue{/color cells/textcolor}\textcolorvalue%
        %
        \toks0=\expandafter{\typesetvalue}%
        \xdef\III{0}%
        \ifnum\III=\pgfplotstablecol\relax%
        \def\addedContent{\pgfplotstablecolname&}%
        \else%
        \def\addedContent{}%
        \fi%
        \xdef\temp{%
          \noexpand\pgfkeysalso{%
            @cell content={%
              \noexpand\cellcolor[rgb]{\pgfmathresult}%
              \noexpand\definecolor{mapped color}{rgb}{\pgfmathresult}%
              \ifx\textcolorvalue\empty%
              \else%
              \noexpand\color{\textcolorvalue}%
              \fi%
              \the\toks0%
            }%
          }%
        }%
        \endgroup%
        \temp%
      }%
    }%
  },
  every head row/.style={
    before row=\toprule&\multicolumn{7}{c|}{Year}\\,%
    after row=\midrule\midrule%
  },
} \begin{table}[tbp]
  \setlength{\tabcolsep}{2.3pt}
  \centering
  \caption{Shows the number of recognition results that were published
    in a specific range of years, modeling a specific \gls{vocabulary} size.
    The top part of the table show all studies jointly, while the
    middle and the bottom part of the table show \gls{isolated} and
    \gls{continuous} studies, respectively. \Eg this table
    reads like: ``After 2015, there were 43 results published tackling
    \glspl{vocabulary} larger than 1000 signs.''}
  \label{tab:vocab_by_year}
  \begin{tabular}{c}
            \rotatebox{90}{All Studies~ ~ ~ ~ ~ ~ ~ }\\
        \end{tabular} 
  \pgfplotstabletypeset[
  /pgfplots/colormap={CM}{color=(white) rgb255=(255,170,0)},
  color cells={min=0,max=52,textcolor=black},
  columns/Vocabulary/.style={
    column type={|r|},
    preproc cell content/.append style={@cell content={$##1$}},
    postproc cell content/.code={}
  },
  every last row/.style={after row=\midrule},
  columns/Global/.style={
    column type={c|},
  },
  columns/<1990/.style={
    column type={c|},
  },
  columns/Motion/.style={column type=c|},
  every cell content/.add={a},
    typeset cell/.append code={%
    \ifnum\pgfplotstablerow<0
      \ifnum\pgfplotstablecol=\pgfplotstablecols
        \pgfkeyssetvalue{/pgfplots/table/@cell content}{$#1$\\}%
      \else
        \ifnum\pgfplotstablecol=1
          \pgfkeyssetvalue{/pgfplots/table/@cell content}{#1&}%
        \else
          \pgfkeyssetvalue{/pgfplots/table/@cell content}{$#1$&}%
        \fi
      \fi
    \fi
  }, 
  ]{histogram-2d-year-vocab.frac.cnt}
  
    \begin{tabular}{c}
            \rotatebox{90}{Isolated Studies~ ~ ~}\\
        \end{tabular}
    \pgfplotstabletypeset[
  /pgfplots/colormap={CM}{color=(white) rgb255=(255,170,0)},
  color cells={min=0,max=52,textcolor=black},
  columns/Vocabulary/.style={
    column type={|r|},
    preproc cell content/.append style={@cell content={$##1$}},
    postproc cell content/.code={}
  },
  every last row/.style={after row=\midrule},
  columns/Global/.style={
    column type={c|},
  },
  columns/<1990/.style={
    column type={c|},
  },
  columns/Motion/.style={column type=c|},
  every cell content/.add={a},
    typeset cell/.append code={%
    \ifnum\pgfplotstablerow<0
      \ifnum\pgfplotstablecol=\pgfplotstablecols
        \pgfkeyssetvalue{/pgfplots/table/@cell content}{$#1$\\}%
      \else
        \ifnum\pgfplotstablecol=1
          \pgfkeyssetvalue{/pgfplots/table/@cell content}{#1&}%
        \else
          \pgfkeyssetvalue{/pgfplots/table/@cell content}{$#1$&}%
        \fi
      \fi
      \fi
    },
  every head row/.style={
    before row=\midrule,%
    after row=\midrule\midrule%
  },
  ]{histogram-2d-year-vocab-iso.frac.cnt}
  
    \begin{tabular}{c}
            \rotatebox{90}{Continuous Studies}\\
        \end{tabular}
   \pgfplotstabletypeset[
  /pgfplots/colormap={CM}{color=(white) rgb255=(255,170,0)},
  color cells={min=0,max=52,textcolor=black},
  columns/Vocabulary/.style={
    column type={|r|},
    preproc cell content/.append style={@cell content={$##1$}},
    postproc cell content/.code={}
  },
  every last row/.style={after row=\bottomrule},
  columns/Global/.style={
    column type={c|},
  },
  columns/<1990/.style={
    column type={c|},
  },
  columns/Motion/.style={column type=c|},
  every cell content/.add={a},
    typeset cell/.append code={%
    \ifnum\pgfplotstablerow<0
      \ifnum\pgfplotstablecol=\pgfplotstablecols
        \pgfkeyssetvalue{/pgfplots/table/@cell content}{$#1$\\}%
      \else
        \ifnum\pgfplotstablecol=1
          \pgfkeyssetvalue{/pgfplots/table/@cell content}{#1&}%
        \else
          \pgfkeyssetvalue{/pgfplots/table/@cell content}{$#1$&}%
        \fi
      \fi
    \fi
  },
    every head row/.style={
    before row=\midrule,%
    after row=\midrule\midrule%
  },
  ]{histogram-2d-year-vocab-cont.frac.cnt}
\end{table}
\pgfplotstableset{
  col sep=space,
}

\subsection{Type of Employed Input Data}
\label{sec:type-input-data}
%
%
Table~\ref{tab:input_data_by_vocabulary_year} shows in the top part of the type of employed
input data across different sizes of modeled \gls{vocabulary}. The
input data refers to the data that is consumed by the recognition
algorithms to extract features from and perform computation.
We can
observe that RGB is the most popular type of input data both for small
and larger scale \gls{vocabulary} ranges. Colored gloves have only ever been
applied to small and medium \gls{vocabulary} tasks and did never get
significant attention.
The lower part of Table~\ref{tab:input_data_by_vocabulary_year} shows the type of employed input
data relative to all results published in the same range of years. We
can see that RGB data attracts most attention since
2005. Depth as input modality became only popular after
the release of the Kinect sensor in 2010. There was one work that
employed depth data before~\cite{fujimura_sign_2006} which had access
to early time-of-flight sensors. Colored gloves got some
traction between 1995 and 2010, which looks like a transition phase
from electronic measuring devices to pure vision based processing.

Table~\ref{tab:intrusive_by_vocab_year} displays the input data aggregated
into the categories `\gls{non-intrusive}' and `\gls{intrusive}'. Intrusiveness refers to the
need to interfere with the recognition subject in order to perform
body pose estimation and general feature extraction. As such, `RGB' and
`Depth' are \gls{non-intrusive} capturing methods, while `Color
Glove', `Electronic Glove' and `Motion Capturing' are \gls{intrusive} techniques. 
As can be seen in Table~\ref{tab:intrusive_by_vocab_year} on the left, \glspl{intrusive} capturing
methods 
can be encountered in
about one quarter of all experiments with a \gls{vocabulary} of up to 500
signs. They are more rare in larger \gls{vocabulary} sizes, possibly due to the
fact that those have mainly been researched after 2010 (compare Table~\ref{tab:vocab_by_year}).
We clearly see
a paradigm shift after 2005, when the formerly dominating \glspl{intrusive} capturing methods were
less and less used and their prevalence decreased from around 70\% to
less than 30\% with a tendency to further reduce over time.

Table~\ref{tab:sign_languages_per_input_data} shows the number of
recognition results per per sign language and employed type of input
data.  We note that experiments recognizing \gls{asl} are clearly
dominated by RGB data. \Gls{csl} has most results using RGB-D (color
with depth) data or just RGB data. Gloves make up a significant number of published results in both
sign languages as well. \Gls{dgs} and most other sign languages focus
mainly on RGB based recognition.

\pgfplotstableset{
  string type,
   col sep=&,
   /color cells/min/.initial=0,
   /color cells/max/.initial=1000,
   /color cells/textcolor/.initial=,
   %
   color cells/.code={%
     \pgfqkeys{/color cells}{#1}%
     \pgfkeysalso{%
       postproc cell content/.code={%
         \pgfplotstablegetelem{\pgfplotstablerow}{sum}\of{histogram-2d-vocab-inputs.frac.sum}
         \def\max{\pgfplotsretval}
         \message{2ndchecking \pgfplotstablecol \pgfplotstablecolname}%
         \begingroup%
         %
         %
         \pgfkeysgetvalue{/pgfplots/table/@preprocessed cell content}\value%
         \pgfmathfloatparsenumber{\value}%
         \pgfmathfloattofixed{\pgfmathresult}%
         \let\value=\pgfmathresult%
         %
         \pgfplotscolormapaccess%
         [\pgfkeysvalueof{/color cells/min}:\max]%
         {\value}%
         {\pgfkeysvalueof{/pgfplots/colormap name}}%
         %
         \pgfkeysgetvalue{/pgfplots/table/@cell content}\typesetvalue%
         \message{row \pgfplotstablerow col \pgfplotstablecol \space matches #1...^^J \pgfplotstablecolname}%
         \pgfkeysgetvalue{/color cells/textcolor}\textcolorvalue%
         %
         \toks0=\expandafter{\typesetvalue}%
         \xdef\III{0}%
         \ifnum\III=\pgfplotstablecol\relax%
         \def\addedContent{\pgfplotstablecolname&}%
         \else%
         \def\addedContent{}%
         \fi%
         \xdef\temp{%
           \noexpand\pgfkeysalso{%
             @cell content={%
               \noexpand\cellcolor[rgb]{\pgfmathresult}%
               \noexpand\definecolor{mapped color}{rgb}{\pgfmathresult}%
               \ifx\textcolorvalue\empty%
               \else%
               \noexpand\color{\textcolorvalue}%
               \fi%
               \the\toks0%
             }%
           }%
         }%
         \endgroup%
         \temp%
       }%
     }%
   },
   every head row/.style={
     before row=\toprule,
     after row=\midrule\midrule%
   },
 } 
 \begin{table}[tbp]
   \centering
   \caption{Shows the fraction in [\%] of published sign language recognition
    results that make use of a specific input data type (\eg
    `RGB', `Depth', \etc) relative to all published
    results that fall in the same modeled \gls{vocabulary} range (top
    part of the table) and that have been published in a similar range
    of years (bottom part of the table)\Eg this table
    reads like: ``86\% of all results with a modeled \gls{vocabulary} above
    1000 signs employ RGB input data. 88\% of all results published
    after 2015 also use depth as input data.''}
   \label{tab:input_data_by_vocabulary_year}
   \setlength{\tabcolsep}{4pt}
   \pgfplotstabletypeset[
   /pgfplots/colormap={CM}{color=(white) rgb255=(255,170,0)},
   color cells={min=0,max=113,textcolor=black},
   columns/Vocabulary/.style={
     column type={|>{\RaggedLeft}p{1.8cm}|},
     preproc cell content/.append style={@cell content={$##1$}},
     postproc cell content/.code={}
   },
   every last row/.style={after row=\bottomrule},
   columns/Mocap/.style={
     column type={c|},
   },
   columns/Mouth/.style={
     column type={c|},
   },
   ]{histogram-2d-vocab-inputs.frac.cnt}
   \pgfplotstableset{
  string type,
  col sep=&,
  /color cells/min/.initial=0,
  /color cells/max/.initial=1000,
  /color cells/textcolor/.initial=,
  %
  color cells/.code={%
    \pgfqkeys{/color cells}{#1}%
    \pgfkeysalso{%
      postproc cell content/.code={%
        \pgfplotstablegetelem{\pgfplotstablerow}{sum}\of{histogram-2d-year-inputs.frac.sum}
        \def\max{\pgfplotsretval}
        \message{2ndchecking \pgfplotstablecol \pgfplotstablecolname}%
        \begingroup%
        %
        %
        \pgfkeysgetvalue{/pgfplots/table/@preprocessed cell content}\value%
        \pgfmathfloatparsenumber{\value}%
        \pgfmathfloattofixed{\pgfmathresult}%
        \let\value=\pgfmathresult%
        %
        \pgfplotscolormapaccess%
        [\pgfkeysvalueof{/color cells/min}:\max]%
        {\value}%
        {\pgfkeysvalueof{/pgfplots/colormap name}}%
        %
        \pgfkeysgetvalue{/pgfplots/table/@cell content}\typesetvalue%
        \message{row \pgfplotstablerow col \pgfplotstablecol \space matches #1...^^J \pgfplotstablecolname}%
        \pgfkeysgetvalue{/color cells/textcolor}\textcolorvalue%
        %
        \toks0=\expandafter{\typesetvalue}%
        \xdef\III{0}%
        \ifnum\III=\pgfplotstablecol\relax%
        \def\addedContent{\pgfplotstablecolname&}%
        \else%
        \def\addedContent{}%
        \fi%
        \xdef\temp{%
          \noexpand\pgfkeysalso{%
            @cell content={%
              \noexpand\cellcolor[rgb]{\pgfmathresult}%
              \noexpand\definecolor{mapped color}{rgb}{\pgfmathresult}%
              \ifx\textcolorvalue\empty%
              \else%
              \noexpand\color{\textcolorvalue}%
              \fi%
              \the\toks0%
            }%
          }%
        }%
        \endgroup%
        \temp%
      }%
    }%
  },
  every head row/.style={
    before row=\toprule,
    after row=\midrule\midrule%
  },
} 
  \pgfplotstabletypeset[
  /pgfplots/colormap={CM}{color=(white) rgb255=(255,170,0)},
  color cells={min=0,max=113,textcolor=black},
  columns/Year/.style={
    column type={|>{\RaggedLeft}p{1.8cm}|},
    preproc cell content/.append style={@cell content={$##1$}},
    postproc cell content/.code={}
  },
  every last row/.style={after row=\bottomrule},
  columns/Mocap/.style={
    column type={c|},
  },
  columns/Mouth/.style={
    column type={c|},
  },
  ]{histogram-2d-year-inputs.frac.cnt}
\end{table}
\pgfplotstableset{
  col sep=space,
}

\pgfplotstableset{
  string type,
  col sep=&,
  /color cells/min/.initial=0,
  /color cells/max/.initial=1000,
  /color cells/textcolor/.initial=,
  %
  color cells/.code={%
    \pgfqkeys{/color cells}{#1}%
    \pgfkeysalso{%
      postproc cell content/.code={%
        \pgfplotstablegetelem{\pgfplotstablerow}{sum}\of{histogram-2d-vocab-intrusive.frac.sum}
        \def\max{\pgfplotsretval}
        \message{2ndchecking \pgfplotstablecol \pgfplotstablecolname}%
        \begingroup%
        %
        %
        \pgfkeysgetvalue{/pgfplots/table/@preprocessed cell content}\value%
        \pgfmathfloatparsenumber{\value}%
        \pgfmathfloattofixed{\pgfmathresult}%
        \let\value=\pgfmathresult%
        %
        \pgfplotscolormapaccess%
        [\pgfkeysvalueof{/color cells/min}:\max]%
        {\value}%
        {\pgfkeysvalueof{/pgfplots/colormap name}}%
        %
        \pgfkeysgetvalue{/pgfplots/table/@cell content}\typesetvalue%
        \message{row \pgfplotstablerow col \pgfplotstablecol \space matches #1...^^J \pgfplotstablecolname}%
        \pgfkeysgetvalue{/color cells/textcolor}\textcolorvalue%
        %
        \toks0=\expandafter{\typesetvalue}%
        \xdef\III{0}%
        \ifnum\III=\pgfplotstablecol\relax%
        \def\addedContent{\pgfplotstablecolname&}%
        \else%
        \def\addedContent{}%
        \fi%
        \xdef\temp{%
          \noexpand\pgfkeysalso{%
            @cell content={%
              \noexpand\cellcolor[rgb]{\pgfmathresult}%
              \noexpand\definecolor{mapped color}{rgb}{\pgfmathresult}%
              \ifx\textcolorvalue\empty%
              \else%
              \noexpand\color{\textcolorvalue}%
              \fi%
              \the\toks0%
            }%
          }%
        }%
        \endgroup%
        \temp%
      }%
    }%
  },
  every head row/.style={
    before row=\toprule,
    after row=\midrule\midrule%
  },
} 
\begin{table}[tbp]
  \centering
  \caption{Shows the fraction in [\%] of published sign language recognition
    results that make use of \glspl{non-intrusive} data input capturing methods (\ie
    `RGB' or `Depth') and those that are \glspl{intrusive} (\ie `Color Glove',
    `Elect. Glove' or `Mocap') relative to all
    published results that fall in the same modeled \gls{vocabulary} range
    (left table) and relative to a year range (right table). \Eg this table
    reads like: ``84\% of all published results with a modeled
    \gls{vocabulary} larger than 1000 signs employ
    \glspl{non-intrusive} input data capturing methods.''}
  \label{tab:intrusive_by_vocab_year}
  \setlength{\tabcolsep}{4pt}
  \pgfplotstabletypeset[
  /pgfplots/colormap={CM}{color=(white) rgb255=(255,170,0)},
  color cells={min=0,max=113,textcolor=black},
  columns/Vocabulary/.style={
    column type={|r|},
    preproc cell content/.append style={@cell content={$##1$}},
    postproc cell content/.code={}
  },
  every last row/.style={after row=\bottomrule},
  columns/Intrusive/.style={
    column type={c|},
  },
  columns/Mouth/.style={
    column type={c|},
  },
  ]{histogram-2d-vocab-intrusive.frac.cnt}
\pgfplotstableset{
  string type,
  col sep=&,
  /color cells/min/.initial=0,
  /color cells/max/.initial=1000,
  /color cells/textcolor/.initial=,
  %
  color cells/.code={%
    \pgfqkeys{/color cells}{#1}%
    \pgfkeysalso{%
      postproc cell content/.code={%
        \pgfplotstablegetelem{\pgfplotstablerow}{sum}\of{histogram-2d-year-intrusive.frac.sum}
        \def\max{\pgfplotsretval}
        \message{2ndchecking \pgfplotstablecol \pgfplotstablecolname}%
        \begingroup%
        %
        %
        \pgfkeysgetvalue{/pgfplots/table/@preprocessed cell content}\value%
        \pgfmathfloatparsenumber{\value}%
        \pgfmathfloattofixed{\pgfmathresult}%
        \let\value=\pgfmathresult%
        %
        \pgfplotscolormapaccess%
        [\pgfkeysvalueof{/color cells/min}:\max]%
        {\value}%
        {\pgfkeysvalueof{/pgfplots/colormap name}}%
        %
        \pgfkeysgetvalue{/pgfplots/table/@cell content}\typesetvalue%
        \message{row \pgfplotstablerow col \pgfplotstablecol \space matches #1...^^J \pgfplotstablecolname}%
        \pgfkeysgetvalue{/color cells/textcolor}\textcolorvalue%
        %
        \toks0=\expandafter{\typesetvalue}%
        \xdef\III{0}%
        \ifnum\III=\pgfplotstablecol\relax%
        \def\addedContent{\pgfplotstablecolname&}%
        \else%
        \def\addedContent{}%
        \fi%
        \xdef\temp{%
          \noexpand\pgfkeysalso{%
            @cell content={%
              \noexpand\cellcolor[rgb]{\pgfmathresult}%
              \noexpand\definecolor{mapped color}{rgb}{\pgfmathresult}%
              \ifx\textcolorvalue\empty%
              \else%
              \noexpand\color{\textcolorvalue}%
              \fi%
              \the\toks0%
            }%
          }%
        }%
        \endgroup%
        \temp%
      }%
    }%
  },
  every head row/.style={
    before row=\toprule,
    after row=\midrule\midrule%
  },
} 
  \pgfplotstabletypeset[
  /pgfplots/colormap={CM}{color=(white) rgb255=(255,170,0)},
  color cells={min=0,max=113,textcolor=black},
  columns/Year/.style={
    column type={|r|},
    preproc cell content/.append style={@cell content={$##1$}},
    postproc cell content/.code={}
  },
  every last row/.style={after row=\bottomrule},
  columns/Intrusive/.style={
    column type={c|},
  },
  columns/Mouth/.style={
    column type={c|},
  },
  ]{histogram-2d-year-intrusive.frac.cnt}
\end{table}
\pgfplotstableset{
  col sep=space,
}

\pgfplotstableset{
  string type,
  col sep=&,
  /color cells/min/.initial=0,
  /color cells/max/.initial=1000,
  /color cells/textcolor/.initial=,  
  color cells/.code={%
    \pgfqkeys{/color cells}{#1}%
    \pgfkeysalso{%
      postproc cell content/.code={%
        \pgfplotstablegetelem{\pgfplotstablerow}{sum}\of{histogram-2d-signlanguage-inputdata.sum}
        \def\max{\pgfplotsretval}
        \message{2ndchecking \pgfplotstablecol \space ::
          \space \pgfplotstablecolname \space ::}%
        \begingroup%
        %
                 %
        \pgfkeysgetvalue{/pgfplots/table/@preprocessed cell content}\value%
        \message{wir sind cell value \value}
        \pgfmathfloatparsenumber{\value}%
        \pgfmathfloattofixed{\pgfmathresult}%
        \let\value=\pgfmathresult%
        %
        \pgfplotscolormapaccess%
        [\pgfkeysvalueof{/color cells/min}:\max]%
        {\value}%
        {\pgfkeysvalueof{/pgfplots/colormap name}}%
        %
        \pgfkeysgetvalue{/pgfplots/table/@cell content}\typesetvalue%
        \message{row \pgfplotstablerow col \pgfplotstablecol \space matches #1...^^J \pgfplotstablecolname}%
        \pgfkeysgetvalue{/color cells/textcolor}\textcolorvalue%
        %
        \toks0=\expandafter{\typesetvalue}%
        \xdef\III{0}%
        \ifnum\III=\pgfplotstablecol\relax%
        \def\addedContent{\pgfplotstablecolname&}%
        \else%
        \def\addedContent{}%
        \fi%
        \xdef\temp{%
          \noexpand\pgfkeysalso{%
            @cell content={%
              \noexpand\cellcolor[rgb]{\pgfmathresult}%
              \noexpand\definecolor{mapped color}{rgb}{\pgfmathresult}%
              \ifx\textcolorvalue\empty%
              \else%
              \noexpand\color{\textcolorvalue}%
              \fi%
              \the\toks0%
            }%
          }%
        }%
        \endgroup%
        \temp%
      }%
    }%
  },
  every head row/.style={
    before row=\toprule,
    after row=\midrule\midrule%
  },
} 
\begin{table}[tbp]
  \setlength{\tabcolsep}{2.3pt}
  \centering
  \caption{Shows the number of published recognition results per sign
    language and type of input data. The sign language abbreviations are mentioned
    in the appendix. The sign languages are ordered by result counts. This table
    reads like: ``99 results were published for \gls{asl} that used RGB input data.''}
  \label{tab:sign_languages_per_input_data}
  \pgfplotstabletypeset[
  /pgfplots/colormap={CM}{color=(white) rgb255=(255,170,0)},
  color cells={min=0,max=113,textcolor=black},
  columns/Input Data/.style={
    column type={|c|},
    postproc cell content/.code={}
  },
  every last row/.style={after row=\bottomrule},
  columns/isl/.style={
    column type={c|},
  },
  columns/Gaze/.style={
    column type={c|},
  },
  columns/Motion/.style={column type=c|},
  typeset cell/.append code={%
    \ifnum\pgfplotstablerow<0
    \ifnum\pgfplotstablecol=\pgfplotstablecols
    \pgfkeyssetvalue{/pgfplots/table/@cell content}{\rotatebox{90}{\acrshort{#1}}\\}%
    \else
    \ifnum\pgfplotstablecol=1
    \pgfkeyssetvalue{/pgfplots/table/@cell content}{#1&}%
    \else
    \pgfkeyssetvalue{/pgfplots/table/@cell content}{\rotatebox{90}{\acrshort{#1}}&}%
    \fi
    \fi
    \fi
  },   
  ]{histogram-2d-signlanguage-inputdata.cnt}
\end{table}
\pgfplotstableset{
  col sep=space,
}

%
%
\subsection{Modeled Sign Language Parameters}
\label{sec:modeled-sign-language-modalities}
%
%
In the previous section, we have looked at what kind of input data is
being employed for sign language recognition studies. Now, we will investigate the
sign language \glspl{parameter} and features that are extracted based
on the input data. Therefore, we tagged which sign language
\glspl{parameter} are covered by the modeled features. We
distinguish manual \glspl{parameter} (\ie hand shape, movement,
location and orientation) and non-manual \glspl{parameter} (\ie head,
mouth, eyes, eye blink, eye brows and eye gaze).
For non-manual \glspl{parameter}, it needs
to be pointed out that we focused on studies that 
explicitly target sign language recognition and also include
non-manuals. There are many works that focus on non-manual marker
recognition for sign language, but these works typically do not model
a sign language recognition problem. 
Additionally, we
track common features that capture a global view of the signers (\ie
body joints, fullframe RGB images covering the full signer, fullframe
depth images and fullframe motion images with optical flow). 
Table~\ref{tab:modality_by_vocabulary} shows the employed sign
language parameters and features relative to all results published
using a similar sign \gls{vocabulary}
(top of the table) and relative to all results published during a
similar time (lower part of the table).
 
We note
that hand shape is the most covered \gls{parameter}, while location and
movement are the next popular \glspl{parameter} across all  \gls{vocabulary}
sizes below 1000 signs. Fullframe features followed by hand shapes
are most frequently encountered in large \gls{vocabulary} tasks beyond 1000
signs. The lower part of Table~\ref{tab:modality_by_vocabulary}
confirms that since 2015 fullframe features have become the most
frequently encountered feature (while being very close to hand shape
features).
Furthermore, it can be noticed that since 2015 hand shape are tackled
by a much larger fraction of published results.  
It needs to be pointed out that while most studies that have been
published after 2015 employ a cropped hand patch as input to their
recognition systems, we tagged that with the hand shape
parameter. However, using deep learning based feature extractors, such
hand inputs may implicitly learn hand posture / 
orientation parameters. Similarly, global input features such as fullframe inputs
may implicitly help to learn location and movement parameters and, to
a lesser degree, all
other parameters as well as the full image comprises all available information.

Table~\ref{tab:manualnonmanual_by_vocabulary} aggregates hand
location, movement, shape and orientation into manual \glspl{parameter}. Head,
mouth, eyes, eye blink, eyebrows and eye gaze are referred to as
non-manual \glspl{parameter}. Body joints, fullframe, depth and motion are all computed
on the full image and hence we call them global features. We can
see that with larger modeled \glspl{vocabulary} the trend goes from manual to
global features (left side of
Table~\ref{tab:manualnonmanual_by_vocabulary}), where the latter
increase from 18\% usage across all
published results with \glspl{vocabulary} of up to 50 signs to 62\% with
large \glspl{vocabulary} above 1000 signs.
The increase of global features may have two reasons:
\begin{enumerate}
\item The availability of body joints and full depth image features
  with the release of the Kinect in 2010.
\item The shift towards deep learning and trend to input fullframes
  instead of manual feature engineering. 
\end{enumerate}
Both hypotheses can be confirmed by looking at the right side of
Table~\ref{tab:manualnonmanual_by_vocabulary}. There, we see that
global features started gaining traction just after 2010 (release of
the Kinect) and also coincides with when deep learning for sign
language took off in 2015.

While for the previous tables each sign language parameter has been looked at
separately and tagged when present, Table~\ref{tab:modalitycombination_by_vocabulary} shows the frequency
of combinations of features over different vocabularies.  Hence, if a study
models two types of \glspl{parameter} their combination will appear in this
table.
Inline with previous results, we see that
fullframe features alone are by far the most popular on large
\gls{vocabulary} ($>1000$
signs) tasks. They are followed by hand shape features and bodyjoints. 
On very small \gls{vocabulary} ($<50$ signs) tasks, a preference on
hand shape features can be noticed.

Table~\ref{tab:sign_languages_per_sign_modality} shows the number
of published results with employed parameters broken down per sign language.
In the top part of the table all studies are reflected, while the
lower part of the table only shows studies with a \gls{vocabulary} of at
least 200 signs.
We see that while \gls{asl}
has the most published results overall, non-manual parameters (\eg
head, mouth or eyes) are most
frequently included in studies on \gls{dgs}. It is also
striking that despite the fact that \gls{csl} is the second most frequently
researched sign language, there is only a single study that includes
non-manual parameters like the face~\cite{zhou_spatialtemporal_2020}. We also note that there are studies on
smaller sign languages such as \gls{krsl} that explicitly focus on
non-manual parameters~\cite{mukushev_evaluation_2020,sabyrov_realtime_2019}.
Eyes and specifically eyebrows have only been tackled in few
studies~\cite{koller_automatic_2016,koller_continuous_2015,koller_deep_2016,mukushev_evaluation_2020,sabyrov_realtime_2019,yang_combination_2011,zhang_multimodality_2016},
while, to the best of our knowledge, no single work has explicitly included eye gaze or eye blinks
for sign language recognition. 
In the lower part of Table~\ref{tab:sign_languages_per_sign_modality}
studies are limited to have at least a \gls{vocabulary} of 200
signs. Besides two \gls{bsl} studies~\cite{albanie_bsl1k_2020,von_agris_recent_2008}, all others are works on
\gls{dgs}, covering the 450 sign \gls{vocabulary} corpus
SIGNUM~\cite{oberdorfer_investigations_2012,von_agris_significance_2008}
and the 1080 sign \gls{vocabulary} corpus
\phoenix~\cite{forster_improving_2013,forster_modality_2013,koller_automatic_2016,koller_continuous_2015,koller_deep_2016,zhou_spatialtemporal_2020}. \cite{zhou_spatialtemporal_2020}
is the first work that uses the face in a deep learning based large
vocabulary task. 

\newcolumntype{R}[1]{>{\raggedleft\hspace{0pt}}p{#1}}
\pgfplotstableset{
  string type,
  col sep=&,
  /color cells/min/.initial=0,
  /color cells/max/.initial=1000,
  /color cells/textcolor/.initial=,
  %
  color cells/.code={%
    \pgfqkeys{/color cells}{#1}%
    \pgfkeysalso{%
      postproc cell content/.code={%
        \pgfplotstablegetelem{\pgfplotstablerow}{sum}\of{histogram-2d-vocab-features.frac.sum}
        \def\max{\pgfplotsretval}
        \message{2ndchecking \pgfplotstablecol \pgfplotstablecolname}%
        \begingroup%
        %
        %
        \pgfkeysgetvalue{/pgfplots/table/@preprocessed cell content}\value%
        \pgfmathfloatparsenumber{\value}%
        \pgfmathfloattofixed{\pgfmathresult}%
        \let\value=\pgfmathresult%
        %
        \pgfplotscolormapaccess%
        [\pgfkeysvalueof{/color cells/min}:\max]%
        {\value}%
        {\pgfkeysvalueof{/pgfplots/colormap name}}%
        %
        \pgfkeysgetvalue{/pgfplots/table/@cell content}\typesetvalue%
        \message{row \pgfplotstablerow col \pgfplotstablecol \space matches #1...^^J \pgfplotstablecolname}%
        \pgfkeysgetvalue{/color cells/textcolor}\textcolorvalue%
        %
        \toks0=\expandafter{\typesetvalue}%
        \xdef\III{0}%
        \ifnum\III=\pgfplotstablecol\relax%
        \def\addedContent{\pgfplotstablecolname&}%
        \else%
        \def\addedContent{}%
        \fi%
        \xdef\temp{%
          \noexpand\pgfkeysalso{%
            @cell content={%
              \noexpand\cellcolor[rgb]{\pgfmathresult}%
              \noexpand\definecolor{mapped color}{rgb}{\pgfmathresult}%
              \ifx\textcolorvalue\empty%
              \else%
              \noexpand\color{\textcolorvalue}%
              \fi%
              \the\toks0%
            }%
          }%
        }%
        \endgroup%
        \temp%
      }%
    }%
  },
  every head row/.style={
    before row=\toprule,
    after row=\midrule\midrule%
  },
} \begin{table}[tbp]
  \setlength{\tabcolsep}{2.3pt}
  \centering
  \caption{Shows the fraction in [\%] of published sign language recognition
    results that make use of a specific sign language parameter (\eg
    `Loc.', `Mov.', \etc) relative to all published
    results that fall in the same \gls{vocabulary} range (top part of
    the table), or in the same range of years (lower part of the table). `Loc.', `Mov.', `Shape' and `Orient.' stand for
    hand location, movement, shape and orientation (manual
    parameters). `Joints' refers to tracked body joint
    locations. `Fullframe' and `Depth' are the full RGB and depth
    image, respectively, while `Motion' unites all types of motion
    estimation on the full image (often optical flow). \Eg this table
    reads like: ``27\% of all results with a modeled \gls{vocabulary} above
    1000 signs include the location modality.''}
  \label{tab:modality_by_vocabulary}
  \pgfplotstabletypeset[
  /pgfplots/colormap={CM}{color=(white) rgb255=(255,170,0)},
  color cells={min=0,max=113,textcolor=black},
  columns/Vocabulary/.style={
    column type={|R{1.9cm}|},
    preproc cell content/.append style={@cell content={$##1$}},
    postproc cell content/.code={}
  },
  every last row/.style={after row=\bottomrule},
  columns/Orient./.style={
    column type={c|},
  },
  columns/Gaze/.style={
    column type={c|},
  },
  columns/Motion/.style={column type=c|},
  every cell content/.add={a},
  ]{histogram-2d-vocab-features.frac.cnt}
\pgfplotstableset{
  string type,
  col sep=&,
  /color cells/min/.initial=0,
  /color cells/max/.initial=1000,
  /color cells/textcolor/.initial=,
  %
  color cells/.code={%
    \pgfqkeys{/color cells}{#1}%
    \pgfkeysalso{%
      postproc cell content/.code={%
        \pgfplotstablegetelem{\pgfplotstablerow}{sum}\of{histogram-2d-year-features.frac.sum}
        \def\max{\pgfplotsretval}
        \message{2ndchecking \pgfplotstablecol \pgfplotstablecolname}%
        \begingroup%
        %
        %
        \pgfkeysgetvalue{/pgfplots/table/@preprocessed cell content}\value%
        \pgfmathfloatparsenumber{\value}%
        \pgfmathfloattofixed{\pgfmathresult}%
        \let\value=\pgfmathresult%
        %
        \pgfplotscolormapaccess%
        [\pgfkeysvalueof{/color cells/min}:\max]%
        {\value}%
        {\pgfkeysvalueof{/pgfplots/colormap name}}%
        %
        \pgfkeysgetvalue{/pgfplots/table/@cell content}\typesetvalue%
        \message{row \pgfplotstablerow col \pgfplotstablecol \space matches #1...^^J \pgfplotstablecolname}%
        \pgfkeysgetvalue{/color cells/textcolor}\textcolorvalue%
        %
        \toks0=\expandafter{\typesetvalue}%
        \xdef\III{0}%
        \ifnum\III=\pgfplotstablecol\relax%
        \def\addedContent{\pgfplotstablecolname&}%
        \else%
        \def\addedContent{}%
        \fi%
        \xdef\temp{%
          \noexpand\pgfkeysalso{%
            @cell content={%
              \noexpand\cellcolor[rgb]{\pgfmathresult}%
              \noexpand\definecolor{mapped color}{rgb}{\pgfmathresult}%
              \ifx\textcolorvalue\empty%
              \else%
              \noexpand\color{\textcolorvalue}%
              \fi%
              \the\toks0%
            }%
          }%
        }%
        \endgroup%
        \temp%
      }%
    }%
  },
  every head row/.style={
    before row=\toprule,
    after row=\midrule\midrule%
  },
}
  \pgfplotstabletypeset[
  /pgfplots/colormap={CM}{color=(white) rgb255=(255,170,0)},
  color cells={min=0,max=113,textcolor=black},
  columns/Year/.style={
    column type={|R{1.9cm}|},
    preproc cell content/.append style={@cell content={$##1$}},
    postproc cell content/.code={}
  },
  every last row/.style={after row=\bottomrule},
  columns/Orient./.style={
    column type={c|},
  },
  columns/Gaze/.style={
    column type={c|},
  },
  columns/Motion/.style={column type=c|},
  ]{histogram-2d-year-features.frac.cnt}
\end{table}
\pgfplotstableset{
  col sep=space,
}

\pgfplotstableset{
  string type,
  col sep=&,
  /color cells/min/.initial=0,
  /color cells/max/.initial=1000,
  /color cells/textcolor/.initial=,
  %
  color cells/.code={%
    \pgfqkeys{/color cells}{#1}%
    \pgfkeysalso{%
      postproc cell content/.code={%
        \pgfplotstablegetelem{\pgfplotstablerow}{sum}\of{histogram-2d-vocab-manualnonmanual.frac.sum}
        \def\max{\pgfplotsretval}
        \message{2ndchecking \pgfplotstablecol \pgfplotstablecolname}%
        \begingroup%
        %
        %
        \pgfkeysgetvalue{/pgfplots/table/@preprocessed cell content}\value%
        \pgfmathfloatparsenumber{\value}%
        \pgfmathfloattofixed{\pgfmathresult}%
        \let\value=\pgfmathresult%
        %
        \pgfplotscolormapaccess%
        [\pgfkeysvalueof{/color cells/min}:\max]%
        {\value}%
        {\pgfkeysvalueof{/pgfplots/colormap name}}%
        %
        \pgfkeysgetvalue{/pgfplots/table/@cell content}\typesetvalue%
        \message{row \pgfplotstablerow col \pgfplotstablecol \space matches #1...^^J \pgfplotstablecolname}%
        \pgfkeysgetvalue{/color cells/textcolor}\textcolorvalue%
        %
        \toks0=\expandafter{\typesetvalue}%
        \xdef\III{0}%
        \ifnum\III=\pgfplotstablecol\relax%
        \def\addedContent{\pgfplotstablecolname&}%
        \else%
        \def\addedContent{}%
        \fi%
        \xdef\temp{%
          \noexpand\pgfkeysalso{%
            @cell content={%
              \noexpand\cellcolor[rgb]{\pgfmathresult}%
              \noexpand\definecolor{mapped color}{rgb}{\pgfmathresult}%
              \ifx\textcolorvalue\empty%
              \else%
              \noexpand\color{\textcolorvalue}%
              \fi%
              \the\toks0%
            }%
          }%
        }%
        \endgroup%
        \temp%
      }%
    }%
  },
  every head row/.style={
    before row=\toprule,
    after row=\midrule\midrule%
  },
} \begin{table}[tbp]
  \setlength{\tabcolsep}{2.3pt}
  \centering
  \caption{Shows the fraction in [\%] of published sign language recognition
    results that employ manual, non-manual or global features relative
    to all published results that fall in the same \gls{vocabulary} range
    (left side) or the same range of years (right side).
    Manual parameters refer to hand location, movement, shape and
    orientation. Non-manual parameters are head, mouth, eyes, eyeblink,
    eyebrow and eyegaze features. Global features refer to body
    joints, fullframe, depth and motion features. \Eg this table
    reads like: ``46\% of all results with a modeled \gls{vocabulary} above
    1000 signs include manual parameters.''}
  \label{tab:manualnonmanual_by_vocabulary}
  \pgfplotstabletypeset[
  /pgfplots/colormap={CM}{color=(white) rgb255=(255,170,0)},
  color cells={min=0,max=113,textcolor=black},
  columns/Vocabulary/.style={
    column type={|r|},
    preproc cell content/.append style={@cell content={$##1$}},
    postproc cell content/.code={}
  },
  every last row/.style={after row=\bottomrule},
  columns/Global/.style={
    column type={c|},
  },
  columns/Gaze/.style={
    column type={c|},
  },
  columns/Motion/.style={column type=c|},
  every cell content/.add={a},
  ]{histogram-2d-vocab-manualnonmanual.frac.cnt}%
\pgfplotstableset{
  string type,
  col sep=&,
  /color cells/min/.initial=0,
  /color cells/max/.initial=1000,
  /color cells/textcolor/.initial=,
  %
  color cells/.code={%
    \pgfqkeys{/color cells}{#1}%
    \pgfkeysalso{%
      postproc cell content/.code={%
        \pgfplotstablegetelem{\pgfplotstablerow}{sum}\of{histogram-2d-year-manualnonmanual.frac.sum}
        \def\max{\pgfplotsretval}
        \message{2ndchecking \pgfplotstablecol \pgfplotstablecolname}%
        \begingroup%
        %
        %
        \pgfkeysgetvalue{/pgfplots/table/@preprocessed cell content}\value%
        \pgfmathfloatparsenumber{\value}%
        \pgfmathfloattofixed{\pgfmathresult}%
        \let\value=\pgfmathresult%
        %
        \pgfplotscolormapaccess%
        [\pgfkeysvalueof{/color cells/min}:\max]%
        {\value}%
        {\pgfkeysvalueof{/pgfplots/colormap name}}%
        %
        \pgfkeysgetvalue{/pgfplots/table/@cell content}\typesetvalue%
        \message{row \pgfplotstablerow col \pgfplotstablecol \space matches #1...^^J \pgfplotstablecolname}%
        \pgfkeysgetvalue{/color cells/textcolor}\textcolorvalue%
        %
        \toks0=\expandafter{\typesetvalue}%
        \xdef\III{0}%
        \ifnum\III=\pgfplotstablecol\relax%
        \def\addedContent{\pgfplotstablecolname&}%
        \else%
        \def\addedContent{}%
        \fi%
        \xdef\temp{%
          \noexpand\pgfkeysalso{%
            @cell content={%
              \noexpand\cellcolor[rgb]{\pgfmathresult}%
              \noexpand\definecolor{mapped color}{rgb}{\pgfmathresult}%
              \ifx\textcolorvalue\empty%
              \else%
              \noexpand\color{\textcolorvalue}%
              \fi%
              \the\toks0%
            }%
          }%
        }%
        \endgroup%
        \temp%
      }%
    }%
  },
  every head row/.style={
    before row=\toprule,
    after row=\midrule\midrule%
  },
}
  \pgfplotstabletypeset[
  /pgfplots/colormap={CM}{color=(white) rgb255=(255,170,0)},
  color cells={min=0,max=113,textcolor=black},
  columns/Year/.style={
    column type={|r|},
    preproc cell content/.append style={@cell content={$##1$}},
    postproc cell content/.code={}
  },
  every last row/.style={after row=\bottomrule},
  columns/Global/.style={
    column type={c|},
  },
  columns/Gaze/.style={
    column type={c|},
  },
  columns/Motion/.style={column type=c|},
  every cell content/.add={a},
  ]{histogram-2d-year-manualnonmanual.frac.cnt}
\end{table}
\pgfplotstableset{
  col sep=space,
}

\pgfplotstableset{
  string type,
  col sep=&,
  /color cells/min/.initial=0,
  /color cells/max/.initial=1000,
  /color cells/textcolor/.initial=,
  %
  color cells/.code={%
    \pgfqkeys{/color cells}{#1}%
    \pgfkeysalso{%
      postproc cell content/.code={%
        \pgfplotstablegetelem{\pgfplotstablerow}{sum}\of{histogram-2d-vocab-features.frac.comb.sum}
        \def\max{\pgfplotsretval}
        \message{2ndchecking \pgfplotstablecol \pgfplotstablecolname}%
        \begingroup%
        %
        %
        \pgfkeysgetvalue{/pgfplots/table/@preprocessed cell content}\value%
        \pgfmathfloatparsenumber{\value}%
        \pgfmathfloattofixed{\pgfmathresult}%
        \let\value=\pgfmathresult%
        %
        \pgfplotscolormapaccess%
        [\pgfkeysvalueof{/color cells/min}:\max]%
        {\value}%
        {\pgfkeysvalueof{/pgfplots/colormap name}}%
        %
        \pgfkeysgetvalue{/pgfplots/table/@cell content}\typesetvalue%
        \message{row \pgfplotstablerow col \pgfplotstablecol \space matches #1...^^J \pgfplotstablecolname}%
        \pgfkeysgetvalue{/color cells/textcolor}\textcolorvalue%
        %
        \toks0=\expandafter{\typesetvalue}%
        \xdef\III{0}%
        \ifnum\III=\pgfplotstablecol\relax%
        \def\addedContent{\pgfplotstablecolname&}%
        \else%
        \def\addedContent{}%
        \fi%
        \xdef\temp{%
          \noexpand\pgfkeysalso{%
            @cell content={%
              \noexpand\cellcolor[rgb]{\pgfmathresult}%
              \noexpand\definecolor{mapped color}{rgb}{\pgfmathresult}%
              \ifx\textcolorvalue\empty%
              \else%
              \noexpand\color{\textcolorvalue}%
              \fi%
              \the\toks0%
            }%
          }%
        }%
        \endgroup%
        \temp%
      }%
    }%
  },
  every head row/.style={
    before row=\toprule&\multicolumn{5}{c|}{Vocabulary}\\,%
    after row=\midrule\midrule%
  },
} \begin{table}[tbp]
  \setlength{\tabcolsep}{2.3pt}
  \centering
  \caption{Shows the 26 most frequently used modality combinations and
    their relative frequency of use as. This is displayed as fraction
    in [\%] of published sign language recognition
    results that make use of the specific combination of sign language parameters relative to all published
    results that fall in the same \gls{vocabulary} range. `Loc.', `Mov.', `Shape' and `Orient.' stand for
    hand location, movement, shape and orientation (manual
    parameters). `Joints' refers to tracked body joint
    locations. `Fullframe' and `Depth' are the full RGB and depth
    image, respectively, while `Motion' unites all types of motion
    estimation on the full image (often optical flow). \Eg this table
    reads like: ``39\% of all results with a modeled \gls{vocabulary} above
    1000 signs rely fully on the fullframe modality, while 7\% rely on
    the hand shape modality.''}
  \label{tab:modalitycombination_by_vocabulary}
  \pgfplotstabletypeset[
  /pgfplots/colormap={CM}{color=(white) rgb255=(255,170,0)},
  color cells={min=0,max=113,textcolor=black},
  columns/Modality Combination/.style={
    column type={|r|},
    preproc cell content/.append style={@cell content={##1}},
    postproc cell content/.code={}
  },
  every last row/.style={after row=\bottomrule},
  columns/Orient./.style={
    column type={c|},
  },
  columns/0-50/.style={
    column type={c|},
  },
  columns/Motion/.style={column type=c|},
  typeset cell/.append code={%
    \ifnum\pgfplotstablerow<0
      \ifnum\pgfplotstablecol=\pgfplotstablecols
        \pgfkeyssetvalue{/pgfplots/table/@cell content}{$#1$\\}%
      \else
        \ifnum\pgfplotstablecol=1
          \pgfkeyssetvalue{/pgfplots/table/@cell content}{#1&}%
        \else
          \pgfkeyssetvalue{/pgfplots/table/@cell content}{$#1$&}%
        \fi
      \fi
    \fi
  }, 
  ]{histogram-2d-vocab-features.frac.comb.cnt}
\end{table}
\pgfplotstableset{
  col sep=space,
}

\newcolumntype{Q}[1]{>{\centering\hspace{-1.8pt}}p{#1}}%
\pgfplotstableset{
  string type,
  col sep=&,
  /color cells/min/.initial=0,
  /color cells/max/.initial=1000,
  /color cells/textcolor/.initial=,
  typeset cell/.append code={%
    \ifnum\pgfplotstablerow<0
    \ifnum\pgfplotstablecol=\pgfplotstablecols
    \pgfkeyssetvalue{/pgfplots/table/@cell content}{\rotatebox{90}{\acrshort{#1}}\\}%
    \else
    \ifnum\pgfplotstablecol=1
    \pgfkeyssetvalue{/pgfplots/table/@cell content}{#1&}%
    \else
    \pgfkeyssetvalue{/pgfplots/table/@cell content}{\rotatebox{90}{\acrshort{#1}}&}%
    \fi
    \fi
    \fi
  },     
  color cells/.code={%
    \pgfqkeys{/color cells}{#1}%
    \pgfkeysalso{%
      postproc cell content/.code={%
        \pgfplotstablegetelem{\pgfplotstablerow}{sum}\of{histogram-2d-signlanguage-modalities.sum}
        \def\max{\pgfplotsretval}
        \message{2ndchecking \pgfplotstablecol \space ::
          \space \pgfplotstablecolname \space ::}%
        \begingroup%
        %
                 %
        \pgfkeysgetvalue{/pgfplots/table/@preprocessed cell content}\value%
        \message{wir sind cell value \value}
        \pgfmathfloatparsenumber{\value}%
        \pgfmathfloattofixed{\pgfmathresult}%
        \let\value=\pgfmathresult%
        %
        \pgfplotscolormapaccess%
        [\pgfkeysvalueof{/color cells/min}:\max]%
        {\value}%
        {\pgfkeysvalueof{/pgfplots/colormap name}}%
        %
        \pgfkeysgetvalue{/pgfplots/table/@cell content}\typesetvalue%
        \message{row \pgfplotstablerow col \pgfplotstablecol \space matches #1...^^J \pgfplotstablecolname}%
        \pgfkeysgetvalue{/color cells/textcolor}\textcolorvalue%
        %
        \toks0=\expandafter{\typesetvalue}%
        \xdef\III{0}%
        \ifnum\III=\pgfplotstablecol\relax%
        \def\addedContent{\pgfplotstablecolname&}%
        \else%
        \def\addedContent{}%
        \fi%
        \xdef\temp{%
          \noexpand\pgfkeysalso{%
            @cell content={%
              \noexpand\cellcolor[rgb]{\pgfmathresult}%
              \noexpand\definecolor{mapped color}{rgb}{\pgfmathresult}%
              \ifx\textcolorvalue\empty%
              \else%
              \noexpand\color{\textcolorvalue}%
              \fi%
              \the\toks0%
            }%
          }%
        }%
        \endgroup%
        \temp%
      }%
    }%
  },
  every head row/.style={
    before row=\toprule,
    after row=\midrule\midrule%
  },
} 
\begin{table}[tbp]
  \setlength{\tabcolsep}{2.3pt}
  \centering
  \caption{Shows the number of published recognition results per sign
    language and employed sign language modality. The top part of the
    table shows
    all studies, while the lower part only shows studies with a
    \gls{vocabulary} of at least 200 signs.  The sign language abbreviations are mentioned
    in the appendix. This table
    reads like: ``There are 62 results published that use the location
    modality in the recognition of \protect\gls{asl}.''}
  \label{tab:sign_languages_per_sign_modality}
  \pgfplotstabletypeset[
  /pgfplots/colormap={CM}{color=(white) rgb255=(255,170,0)},
  color cells={min=0,max=113,textcolor=black},
  columns/Modalities/.style={
    column type={Q{0.3cm}|c|},
    preproc cell content/.append style={@cell content={&##1}},
    postproc cell content/.code={}
  },
  every head row/.style={before row=\toprule,
  after row=\cmidrule{1-32}\morecmidrules\cmidrule{1-32}},
  every last row/.style={after row=\cmidrule{1-32}\morecmidrules\cmidrule{1-32}},
  columns/isl/.style={
    column type={c|},
  },
  columns/Gaze/.style={
    column type={c|},
  },
  columns/Motion/.style={column type=c|},
  rows/Gaze/.style={
    column type={c|},
  },
  every column/.style={column type={Q{0.29cm}}},
     typeset cell/.append code={%
    \ifnum\pgfplotstablerow<0
    \ifnum\pgfplotstablecol=\pgfplotstablecols
    \pgfkeyssetvalue{/pgfplots/table/@cell content}{\rotatebox{90}{\acrshort{#1}}\\}%
    \else
    \ifnum\pgfplotstablecol=1
    \pgfkeyssetvalue{/pgfplots/table/@cell content}{&#1&}%
    \else
    \pgfkeyssetvalue{/pgfplots/table/@cell content}{\rotatebox{90}{\acrshort{#1}}&}%
    \fi
    \fi
    \fi
    \ifnum\pgfplotstablerow=0
    \ifnum\pgfplotstablecol=1
    \pgfkeyssetvalue{/pgfplots/table/@cell
      content}{\multirow{14}{*}{\rotatebox{90}{All Studies}}#1&}%
    \fi
    \fi
     \ifnum\pgfplotstablerow=3
     \ifnum\pgfplotstablecol=\pgfplotstablecols
       \pgfkeyssetvalue{/pgfplots/table/@cell content}{#1\\\cmidrule{2-32}}%
     \fi
     \fi
     \ifnum\pgfplotstablerow=9
     \ifnum\pgfplotstablecol=\pgfplotstablecols
       \pgfkeyssetvalue{/pgfplots/table/@cell content}{#1\\\cmidrule{2-32}}%
     \fi
     \fi
  }, 
  ]{histogram-2d-signlanguage-modalities.cnt}
\pgfplotstableset{
  string type,
  col sep=&,
  /color cells/min/.initial=0,
  /color cells/max/.initial=1000,
  /color cells/textcolor/.initial=,
  typeset cell/.append code={%
    \ifnum\pgfplotstablerow<0
    \ifnum\pgfplotstablecol=\pgfplotstablecols
    \pgfkeyssetvalue{/pgfplots/table/@cell content}{\rotatebox{90}{\acrshort{#1}}\\}%
    \else
    \ifnum\pgfplotstablecol=1
    \pgfkeyssetvalue{/pgfplots/table/@cell content}{#1&}%
    \else
    \pgfkeyssetvalue{/pgfplots/table/@cell content}{\rotatebox{90}{\acrshort{#1}}&}%
    \fi
    \fi
    \fi
  },     
  color cells/.code={%
    \pgfqkeys{/color cells}{#1}%
    \pgfkeysalso{%
      postproc cell content/.code={%
        \pgfplotstablegetelem{\pgfplotstablerow}{sum}\of{histogram-2d-signlanguage-modalities-over200.sum}
        \def\max{\pgfplotsretval}
        \message{2ndchecking \pgfplotstablecol \space ::
          \space \pgfplotstablecolname \space ::}%
        \begingroup%
        %
                 %
        \pgfkeysgetvalue{/pgfplots/table/@preprocessed cell content}\value%
        \message{wir sind cell value \value}
        \pgfmathfloatparsenumber{\value}%
        \pgfmathfloattofixed{\pgfmathresult}%
        \let\value=\pgfmathresult%
        %
        \pgfplotscolormapaccess%
        [\pgfkeysvalueof{/color cells/min}:\max]%
        {\value}%
        {\pgfkeysvalueof{/pgfplots/colormap name}}%
        %
        \pgfkeysgetvalue{/pgfplots/table/@cell content}\typesetvalue%
        \message{row \pgfplotstablerow col \pgfplotstablecol \space matches #1...^^J \pgfplotstablecolname}%
        \pgfkeysgetvalue{/color cells/textcolor}\textcolorvalue%
        %
        \toks0=\expandafter{\typesetvalue}%
        \xdef\III{0}%
        \ifnum\III=\pgfplotstablecol\relax%
        \def\addedContent{\pgfplotstablecolname&}%
        \else%
        \def\addedContent{}%
        \fi%
        \xdef\temp{%
          \noexpand\pgfkeysalso{%
            @cell content={%
              \noexpand\cellcolor[rgb]{\pgfmathresult}%
              \noexpand\definecolor{mapped color}{rgb}{\pgfmathresult}%
              \ifx\textcolorvalue\empty%
              \else%
              \noexpand\color{\textcolorvalue}%
              \fi%
              \the\toks0%
            }%
          }%
        }%
        \endgroup%
        \temp%
      }%
    }%
  },
  every head row/.style={
  },
} 
  \pgfplotstabletypeset[
  /pgfplots/colormap={CM}{color=(white) rgb255=(255,170,0)},
  color cells={min=0,max=113,textcolor=black},
  columns/Modalities/.style={
    column type={Q{0.3cm}|c|},
    preproc cell content/.append style={@cell content={&##1}},
    postproc cell content/.code={}
  },
  every last row/.style={after row=\bottomrule},
  columns/isl/.style={
    column type={c|},
  },
  columns/Gaze/.style={
    column type={c|},
  },
  columns/Motion/.style={column type=c|},
  rows/Gaze/.style={
    column type={c|},
  },
  every column/.style={column type={Q{0.29cm}}},
     typeset cell/.append code={%
    \ifnum\pgfplotstablerow<0
    \pgfkeyssetvalue{/pgfplots/table/@cell content}{}%
    \fi
    \ifnum\pgfplotstablerow=0
    \ifnum\pgfplotstablecol=1
    \pgfkeyssetvalue{/pgfplots/table/@cell
      content}{\multirow{14}{*}{\rotatebox{90}{Studies with
          \protect\gls{vocabulary} $\geq 200$}}#1&}%
    \fi
    \fi
         \ifnum\pgfplotstablerow=3
     \ifnum\pgfplotstablecol=\pgfplotstablecols
       \pgfkeyssetvalue{/pgfplots/table/@cell content}{#1\\\cmidrule{2-32}}%
     \fi
     \fi
     \ifnum\pgfplotstablerow=9
     \ifnum\pgfplotstablecol=\pgfplotstablecols
       \pgfkeyssetvalue{/pgfplots/table/@cell content}{#1\\\cmidrule{2-32}}%
     \fi
     \fi
  }, 
  ]{histogram-2d-signlanguage-modalities-over200.cnt}
\end{table}
\pgfplotstableset{
  col sep=space,
}

%
%
\subsection{Analysis by Sign Language}
\label{sec:before-and-after-2015}
%
%
Table~\ref{tab:sign_languages_per_year} and
Table~\ref{tab:sign_languages_per_vocabulary} show the number of
published recognition results per sign language over time and per
modeled \gls{vocabulary} range, respectively. \Gls{asl} has usually been the sign language with the most results
published. However, we see in Table~\ref{tab:sign_languages_per_vocabulary}
that this is only true for \glspl{vocabulary} below 200 signs. On larger
\glspl{vocabulary} \gls{csl} is leading and on \glspl{vocabulary} above 1000 signs
\gls{dgs} has significantly more research
published. Table~\ref{tab:sign_languages_per_vocabulary} further
reveals that it is \phoenix with a \gls{vocabulary} of over 1000 signs that
represents the only resource for large-scale \gls{continuous} sign language world wide.

This can partly be explained by the public availability of sign
language data sets, which represent after all still extremely low
resource languages. However, there are corpora available
for \gls{asl} that have larger \glspl{vocabulary} (\eg
ASLLRP~\cite{neidle_new_2012}). It seems there is necessity for the
corpora to be packaged for reproducible sign language recognition
research. Fixed partitions into train, development and test and an
easily accessible download method are required. Also, the licenses
under which the corpora are provided may impact dissemination.

\newcolumntype{Q}[1]{>{\centering\hspace{-1.5pt}}p{#1}}%
\def\tablecolwidth{0.3cm}
\pgfplotstableset{
  string type,
  col sep=&,
  /color cells/min/.initial=0,
  /color cells/max/.initial=1000,
  /color cells/textcolor/.initial=,    
  color cells/.code={%
    \pgfqkeys{/color cells}{#1}%
    \pgfkeysalso{%
      postproc cell content/.code={%
        %
        \message{2ndchecking \pgfplotstablecol \space ::
          \space \pgfplotstablecolname \space ::}%
        \begingroup%
        %
        %
        \pgfkeysgetvalue{/pgfplots/table/@preprocessed cell content}\value%
        \message{wir sind cell value \value}
        \pgfmathfloatparsenumber{\value}%
        \pgfmathfloattofixed{\pgfmathresult}%
        \let\value=\pgfmathresult%
        %
        \pgfplotscolormapaccess%
        [\pgfkeysvalueof{/color cells/min}:\pgfkeysvalueof{/color cells/max}]%
        {\value}%
        {\pgfkeysvalueof{/pgfplots/colormap name}}%
        %
        \pgfkeysgetvalue{/pgfplots/table/@cell content}\typesetvalue%
        \message{row \pgfplotstablerow col \pgfplotstablecol \space matches #1...^^J \pgfplotstablecolname}%
        \pgfkeysgetvalue{/color cells/textcolor}\textcolorvalue%
        %
        \toks0=\expandafter{\typesetvalue}%
        \xdef\III{0}%
        \ifnum\III=\pgfplotstablecol\relax%
        \def\addedContent{\pgfplotstablecolname&}%
        \else%
        \def\addedContent{}%
        \fi%
        \xdef\temp{%
          \noexpand\pgfkeysalso{%
            @cell content={%
              \noexpand\cellcolor[rgb]{\pgfmathresult}%
              \noexpand\definecolor{mapped color}{rgb}{\pgfmathresult}%
              \ifx\textcolorvalue\empty%
              \else%
              \noexpand\color{\textcolorvalue}%
              \fi%
              \the\toks0%
            }%
          }%
        }%
        \endgroup%
        \temp%
      }%
    }%
  },
  every head row/.style={
    after row=\cmidrule{2-32}\morecmidrules\cmidrule{2-32}%
  },
} 
\begin{table}[tbp]
  \setlength{\tabcolsep}{-6.3pt}
  \centering
  \caption{Shows the number of published recognition results per sign
    language and year. The sign language abbreviations are mentioned
    in the appendix.  This table
    reads like: ``There are 50 results published after 2015 that use \protect\gls{asl}.''}
  \label{tab:sign_languages_per_year}
\setlength{\tabcolsep}{2.3pt}%
\pgfplotstabletypeset[
  /pgfplots/colormap={CM}{color=(white) rgb255=(255,170,0)},
  color cells={min=0,max=60,textcolor=black},
  every column/.style={column type={Q{\tablecolwidth}},},%
  every head row/.style={before row=\toprule,
  after row=\cmidrule{2-32}\morecmidrules\cmidrule{2-32}},
  columns/Year/.style={
    column type={c|c|},
    preproc cell content/.append style={@cell content={&$##1$}},
    postproc cell content/.code={}
  },
  columns/isl/.style={
    column type={c|},
  },
  columns/Gaze/.style={
    column type={c|},
  },
  columns/Motion/.style={column type=c|},
    typeset cell/.append code={%
    \ifnum\pgfplotstablerow<0
    \ifnum\pgfplotstablecol=\pgfplotstablecols
    \pgfkeyssetvalue{/pgfplots/table/@cell content}{\rotatebox{90}{\acrshort{#1}}\\}%
    \else
    \ifnum\pgfplotstablecol=1
    \pgfkeyssetvalue{/pgfplots/table/@cell content}{&#1&}%
    \else
    \pgfkeyssetvalue{/pgfplots/table/@cell content}{\rotatebox{90}{\acrshort{#1}}&}%
    \fi
    \fi
    \fi
    \ifnum\pgfplotstablerow=0
    \ifnum\pgfplotstablecol=1
    \pgfkeyssetvalue{/pgfplots/table/@cell
      content}{\multirow{7}{*}{\rotatebox{90}{All Studies}}#1&}%
    \fi
    \fi
  }, 
  ]{histogram-2d-signlanguage-year.cnt}
    \pgfplotstabletypeset[
  /pgfplots/colormap={CM}{color=(white) rgb255=(255,170,0)},
  color cells={min=0,max=60,textcolor=black},
  every column/.style={column type={Q{\tablecolwidth}},},%
  every head row/.style={after row=\cmidrule{2-32}\morecmidrules\cmidrule{2-32}},
  columns/Year/.style={
    column type={c|c|},
    preproc cell content/.append style={@cell content={&$##1$}},
    postproc cell content/.code={}
  },
  columns/isl/.style={
    column type={c|},
  },
  columns/Gaze/.style={
    column type={c|},
  },
  columns/Motion/.style={column type=c|},
  typeset cell/.append code={%
            \ifnum\pgfplotstablerow<0
    \pgfkeyssetvalue{/pgfplots/table/@cell content}{}%
    \fi
    \ifnum\pgfplotstablerow=0
    \ifnum\pgfplotstablecol=1
    \pgfkeyssetvalue{/pgfplots/table/@cell
      content}{\multirow{7}{*}{\rotatebox{90}{Isolated Studies}}#1&}%
    \fi
    \fi
  }, 
  ]{histogram-2d-signlanguage-year-iso.cnt}
    \pgfplotstabletypeset[
  /pgfplots/colormap={CM}{color=(white) rgb255=(255,170,0)},
  color cells={min=0,max=60,textcolor=black},
  every column/.style={column type={Q{\tablecolwidth}},},%
  every head row/.style={after row=\cmidrule{2-32}\morecmidrules\cmidrule{2-32}},
  columns/Year/.style={
    column type={c|c|},
    preproc cell content/.append style={@cell content={&$##1$}},
    postproc cell content/.code={}
  },
  every last row/.style={after row=\bottomrule},
  columns/isl/.style={
    column type={c|},
  },
  columns/Gaze/.style={
    column type={c|},
  },
  columns/Motion/.style={column type=c|},
    columns/Motion/.style={column type=c|},
    typeset cell/.append code={%
              \ifnum\pgfplotstablerow<0
    \pgfkeyssetvalue{/pgfplots/table/@cell content}{}%
    \fi
    \ifnum\pgfplotstablerow=0
    \ifnum\pgfplotstablecol=1
    \pgfkeyssetvalue{/pgfplots/table/@cell
      content}{\multirow{7}{*}{\rotatebox{90}{Continuous Studies}}#1&}%
    \fi
    \fi
  }, 
  ]{histogram-2d-signlanguage-year-cont.cnt}
\end{table}
\pgfplotstableset{
  col sep=space,
}

\pgfplotstableset{
  string type,
  col sep=&,
  /color cells/min/.initial=0,
  /color cells/max/.initial=1000,
  /color cells/textcolor/.initial=,
  typeset cell/.append code={%
    \ifnum\pgfplotstablerow<0
    \ifnum\pgfplotstablecol=\pgfplotstablecols
    \pgfkeyssetvalue{/pgfplots/table/@cell content}{\rotatebox{90}{\acrshort{#1}}\\}%
    \else
    \ifnum\pgfplotstablecol=1
    \pgfkeyssetvalue{/pgfplots/table/@cell content}{#1&}%
    \else
    \pgfkeyssetvalue{/pgfplots/table/@cell content}{\rotatebox{90}{\acrshort{#1}}&}%
    \fi
    \fi
    \fi
  },     
  color cells/.code={%
    \pgfqkeys{/color cells}{#1}%
    \pgfkeysalso{%
      postproc cell content/.code={%
        %
        \message{2ndchecking \pgfplotstablecol \space ::
          \space \pgfplotstablecolname \space ::}%
        \begingroup%
        %
                 %
        \pgfkeysgetvalue{/pgfplots/table/@preprocessed cell content}\value%
        \message{wir sind cell value \value}
        \pgfmathfloatparsenumber{\value}%
        \pgfmathfloattofixed{\pgfmathresult}%
        \let\value=\pgfmathresult%
        %
        \pgfplotscolormapaccess%
        [\pgfkeysvalueof{/color cells/min}:\pgfkeysvalueof{/color cells/max}]%
        {\value}%
        {\pgfkeysvalueof{/pgfplots/colormap name}}%
        %
        \pgfkeysgetvalue{/pgfplots/table/@cell content}\typesetvalue%
        \message{row \pgfplotstablerow col \pgfplotstablecol \space matches #1...^^J \pgfplotstablecolname}%
        \pgfkeysgetvalue{/color cells/textcolor}\textcolorvalue%
        %
        \toks0=\expandafter{\typesetvalue}%
        \xdef\III{0}%
        \ifnum\III=\pgfplotstablecol\relax%
        \def\addedContent{\pgfplotstablecolname&}%
        \else%
        \def\addedContent{}%
        \fi%
        \xdef\temp{%
          \noexpand\pgfkeysalso{%
            @cell content={%
              \noexpand\cellcolor[rgb]{\pgfmathresult}%
              \noexpand\definecolor{mapped color}{rgb}{\pgfmathresult}%
              \ifx\textcolorvalue\empty%
              \else%
              \noexpand\color{\textcolorvalue}%
              \fi%
              \the\toks0%
            }%
          }%
        }%
        \endgroup%
        \temp%
      }%
    }%
  },
  every head row/.style={
    after row=\cmidrule{2-32}\morecmidrules\cmidrule{2-32}%
  },
} 
\begin{table}[tbp]
  \setlength{\tabcolsep}{2.3pt}
  \centering
  \caption{Shows the number of published recognition results per sign
    language and modeled vocabulary. The sign language abbreviations are mentioned
    in the appendix. This table
    reads like: ``There are 6 recognition results of \protect\gls{asl}
    published.''}
  \label{tab:sign_languages_per_vocabulary}
  \pgfplotstabletypeset[
  /pgfplots/colormap={CM}{color=(white) rgb255=(255,170,0)},
  color cells={min=0,max=60,textcolor=black},
  every column/.style={column type={Q{\tablecolwidth}},},%
  columns/Vocabulary/.style={
    column type={c|c|},
    preproc cell content/.append style={@cell content={&$##1$}},
    postproc cell content/.code={}
  },%
  every head row/.style={before row=\toprule,
  after row=\cmidrule{2-32}\morecmidrules\cmidrule{2-32}},
  columns/isl/.style={
    column type={c|},
  },
  columns/Gaze/.style={
    column type={c|},
  },
  columns/Motion/.style={column type=c|},
      typeset cell/.append code={%
    \ifnum\pgfplotstablerow<0
    \ifnum\pgfplotstablecol=\pgfplotstablecols
    \pgfkeyssetvalue{/pgfplots/table/@cell content}{\rotatebox{90}{\acrshort{#1}}\\}%
    \else
    \ifnum\pgfplotstablecol=1
    \pgfkeyssetvalue{/pgfplots/table/@cell content}{&#1&}%
    \else
    \pgfkeyssetvalue{/pgfplots/table/@cell content}{\rotatebox{90}{\acrshort{#1}}&}%
    \fi
    \fi
    \fi
    \ifnum\pgfplotstablerow=0
    \ifnum\pgfplotstablecol=1
    \pgfkeyssetvalue{/pgfplots/table/@cell content}{\multirow{5}{*}{\rotatebox{90}{All Studies}}#1&}%
    \fi
    \fi
  },
  ]{histogram-2d-signlanguage-vocab.cnt}
    \pgfplotstabletypeset[
  /pgfplots/colormap={CM}{color=(white) rgb255=(255,170,0)},
  color cells={min=0,max=60,textcolor=black},
  every column/.style={column type={Q{\tablecolwidth}},},%
  columns/Vocabulary/.style={
    column type={c|c|},
    preproc cell content/.append style={@cell content={&$##1$}},
    postproc cell content/.code={}
  },
  columns/isl/.style={
    column type={c|},
  },
  columns/Gaze/.style={
    column type={c|},
  },
  columns/Motion/.style={column type=c|},
  typeset cell/.append code={%
    \ifnum\pgfplotstablerow<0
    \pgfkeyssetvalue{/pgfplots/table/@cell content}{}%
    \fi
    \ifnum\pgfplotstablerow=0
    \ifnum\pgfplotstablecol=1
    \pgfkeyssetvalue{/pgfplots/table/@cell content}{\multirow{5}{*}{\rotatebox{90}{Isolated}}#1&}%
    \fi
    \fi
  },
  ]{histogram-2d-signlanguage-vocab-iso.cnt}
    \pgfplotstabletypeset[
  /pgfplots/colormap={CM}{color=(white) rgb255=(255,170,0)},
  color cells={min=0,max=60,textcolor=black},
  every column/.style={column type={Q{\tablecolwidth}},},%
  columns/Vocabulary/.style={
    column type={c|c|},
    preproc cell content/.append style={@cell content={&$##1$}},
    postproc cell content/.code={}
  },
  every last row/.style={after row=\bottomrule},
  columns/isl/.style={
    column type={c|},
  },
  columns/Gaze/.style={
    column type={c|},
  },
  columns/Motion/.style={column type=c|},
  typeset cell/.append code={%
        \ifnum\pgfplotstablerow<0
    \pgfkeyssetvalue{/pgfplots/table/@cell content}{}%
    \fi
    \ifnum\pgfplotstablerow=0
    \ifnum\pgfplotstablecol=1
    \pgfkeyssetvalue{/pgfplots/table/@cell content}{\multirow{5}{*}{\rotatebox{90}{Continuous}}#1&}%
    \fi
    \fi
  },
  ]{histogram-2d-signlanguage-vocab-cont.cnt}
\end{table}
\pgfplotstableset{
  col sep=space,
}

%
%
\subsection{Change of Continuous Recognition Landscape After 2015}
\label{sec:before-and-after-2015}
%
%

Figure~\ref{fig:number_of_results_and_modeled_vocabulary_before_after_2015}
shows counts of published \gls{continuous} sign language recognition
experiments and modeled \glspl{vocabulary} before and after 2015. Prior
2015, there were 80 results published, while after 2015 66
results can be found. We
note that prior 2015 most studies model different
\gls{vocabulary} sizes, while after 2015 there is a large peak of close to
30 published results at a 1080
\gls{vocabulary} and a smaller peak at a \gls{vocabulary} size of 178.

Since 2015, the sign language recognition community is focusing more on
benchmark data sets, which explains these characteristics. \phoenix 2014~\cite{koller_continuous_2015} has a \gls{vocabulary} of 1080 and the
CSL corpus~\cite{huang_videobased_2018} covers 178 signs. In the
following section, we will provide a deep analysis of the research
studies that compared their work on the \phoenixs corpus.

\pgfplotstableread[col sep=comma]{histogram-signvocab-count-cont-prior2015.cnt}{\priorvocab}
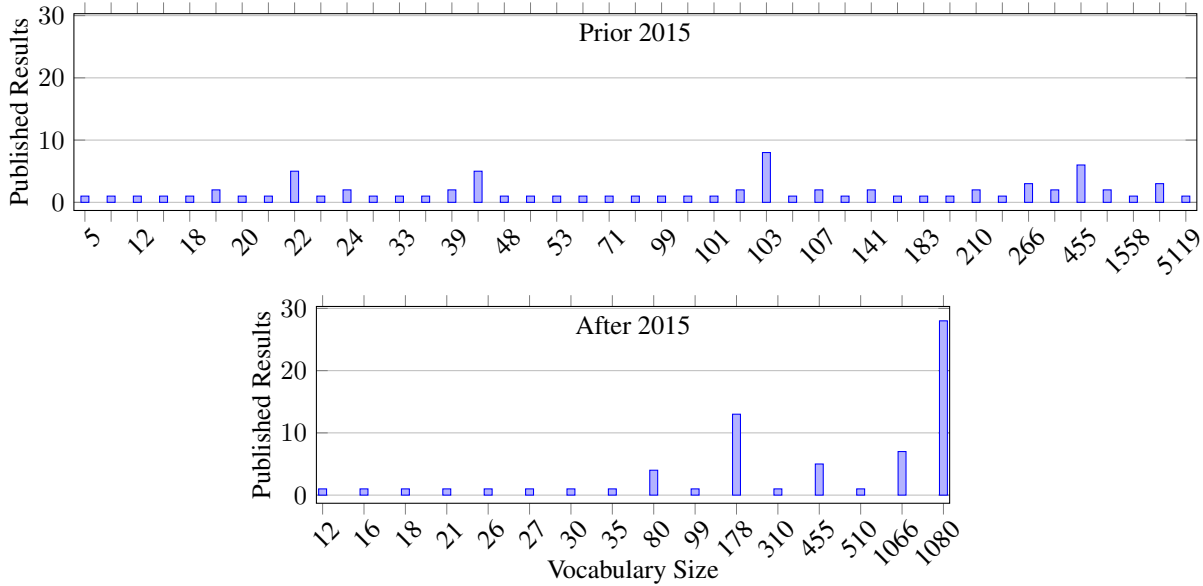
\begin{figure}[htpb]
  \centering
  \begin{tikzpicture}
    \begin{axis}[
      ybar,
      title style={yshift=-0.7cm},
      title={Prior 2015},
      width=\textwidth,
      height=4.2cm,
      bar width=3,
      ymin=-1,
      enlargelimits=0.01,
      xmin=0,
      ymax=30,
      ymajorgrids,
      xtick=data,
      xticklabels from table={\priorvocab}{[index] 0},
      xtick distance=1,
      x tick label style={rotate=45,anchor=east,yshift=-6,xshift=4},
      ylabel={Published Results},
      ylabel style={yshift=-0.5cm},
      xlabel style={yshift=-0.1cm},
      ]
      \addplot table [col sep = comma,
      x expr = \coordindex,
      y = count] {\priorvocab};
    \end{axis}
  \end{tikzpicture}
  \pgfplotstableread[col sep=comma]{histogram-signvocab-count-cont-after2015.cnt}{\aftervocab}
  \begin{tikzpicture}
    \begin{axis}[
      ybar,
      title style={yshift=-0.7cm},
      title={After 2015},
      width=10cm,
      height=4.2cm,
      bar width=3,
      ymin=-1,
      enlargelimits=0.01,
      xmin=0,
      ymax=30,
      ymajorgrids,
      xtick=data,
      xticklabels from table={\aftervocab}{[index] 0},
      xtick distance=1,
      x tick label style={rotate=45,anchor=east,yshift=-6,xshift=4},
      xlabel={Vocabulary Size},
      ylabel={Published Results},
      ylabel style={yshift=-0.5cm},
      xlabel style={yshift=-0.1cm},
      ]
      \addplot table [col sep = comma,
      x expr = \coordindex,
      y = count] {\aftervocab};
    \end{axis}
  \end{tikzpicture}
  \caption{Showing the number of published \gls{continuous} sign language recognition results per
    modeled \gls{vocabulary} (prior to 2015 on the top and 2015-2020 on the
    bottom plot). This allows to see that after 2015 researcher have
    started to focus on few benchmark data sets.
    }
  \label{fig:number_of_results_and_modeled_vocabulary_before_after_2015}
\end{figure}

\section{Analysis of \phoenixs 2014 Benchmark Papers}
\label{sec:phoenix-2014-analysis}

Table~\ref{tab:reported_results_on_phoenix_1} and
Table~\ref{tab:reported_results_on_phoenix_2} present, to the best of
our knowledge, all known
results on the \phoenix 2014 \gls{continuous} sign language
recognition benchmark that have been published as of June 2020.
Table~\ref{tab:reported_results_on_phoenix_1}
provides information on the employed features, the chosen neural
architecture and the achieved
results, while Table~\ref{tab:reported_results_on_phoenix_2} shows what kind
of data augmentation was used, if
iterative training was employed and what training losses were part of
the optimization.
%
Iterative training refers to an \gls{em} like training procedure where
a trained model is used to create pseudo labels on the training data
which will then be used to train a part or the full recognition
network. Inspired by \gls{em} training in \gls{gmm} \gls{hmm} systems,
this way of training was first proposed in
\cite{koller_automatic_2016}. It was then adopted by many teams as can be seen in
Table~\ref{tab:reported_results_on_phoenix_2}. Besides
\cite{cheng_fully_2020}, all best performing approaches on \phoenixs
with a \glspl{wer} below 27.0\% make use of iterative training
procedures. In many works it is described to help overcome vanishing
gradients issues when training deep \gls{cnn} architectures that are
succeeded by \gls{blstm} layers~\cite{zhou_spatialtemporal_2020,papastratis_continuous_2020,cui_deep_2019}.

The employed losses are very diverse, as can be seen in
Table~\ref{tab:reported_results_on_phoenix_2}. However, most networks
that achieve below 30.0\% \gls{wer} are trained with
\gls{ce} loss and also use \gls{ctc} loss. Additionally, a variety of
different loss terms are reported ranging from \gls{kl} divergence,
over squared error to smooth-L1 loss and others.

Table~\ref{tab:reported_results_on_phoenix_1} shows that
\cite{cui_recurrent_2017} first suggested the use of 2D convolutions
followed by 1D convolutions on
\phoenixs. Later, \cite{tran_closer_2018} did a detailed analysis for
action recognition. All best performing approaches on \phoenixs
with \glspl{wer} below 25.0\% employ 2D+1D convolutions~\cite{cheng_fully_2020,cui_deep_2019,papastratis_continuous_2020,zhou_spatialtemporal_2020}.

While the early works on \phoenixs all relied on tracked and cropped
hand shape features \cite{koller_resign_2017} first proposed to
train the \glspl{cnn} directly on the fullframe input image. This
trend continues and all recent studies rely on this feature (\eg \cite{adaloglou_comprehensive_2020,cheng_fully_2020,papastratis_continuous_2020,zhou_selfattentionbased_2020,zhou_spatialtemporal_2020}).

In terms of data augmentation, the most popular choice seems to be
random cropping, followed by temporal scaling (re-sampling or random
frame drop) as can be seen in
Table~\ref{tab:reported_results_on_phoenix_2}. However, many papers do
not specify any augmentation methods leaving the reader without a
clear understanding of what happens. While augmentation certainly has significant impact on
the results, no study has yet analyzed the effect of the various
augmentation options.

Finally, it needs to be pointed out that all of the works that
compared their performance on \phoenixs were based on whole sign units
during inference. Only two works made use of subunits, \ie to additionally guide
the alignment process~\cite{koller_weakly_2019} or in a pretraining stage~\cite{borg_phonologicallymeaningful_2020}.

\newcommand{\stdsize}{\footnotesize}
\def\colorModel{rgb} 
\newcommand\ColCell[1]{%
\pgfmathsetmacro\min{20.7}%
\pgfmathsetmacro\max{55}%
\pgfmathsetmacro\compA{1}
\pgfmathsetmacro\compB{170/255+(1-170/255)*(1-(#1-\min)/(\max-\min))}
\pgfmathsetmacro\compC{1-(#1-\min)/(\max-\min)}
\edef\x{\noexpand\cellcolor[\colorModel]{\compA,\compB,\compC}}%
\x%
#1} 
\newcolumntype{E}{>{\stdsize\collectcell\ColCell}p{0.5cm}<{\endcollectcell}} 

\newcolumntype{F}{>{\collectcell\ColCell\stdsize}m{1.5cm}<{\endcollectcell}}  

\newcolumntype{L}[1]{>{\RaggedRight\hspace{0pt}\scriptsize}p{#1}}
\newcolumntype{Q}[1]{>{\centering\stdsize}p{#1}}
\newcolumntype{P}[1]{>{\stdsize}p{#1}}
\newcolumntype{C}{>{\stdsize}{c}}
\newcolumntype{S}{>{\scriptsize}{c}}
\newcolumntype{R}{>{\stdsize}{r}}
\begin{center}
  \setlength{\tabcolsep}{1.pt}
  { 
    \begin{longtable}{|L{2cm}|SS|CCCC|CCCC|CCC|CCCCCC|EE|}
     \caption{ The table covers (to the best of our knowledge) all
       published sign language recognition works  until mid 2020 that
       report results on the RWTH-Phoenix Weather 2014
       \protect\cite{koller_continuous_2015} task. The works are
       ordered by year and by \protect\gls{wer}. It allows to compare the
       type of employed features (manual, non-manual and fullframe
       features), the employed neural architectures and the achieved
       \gls{wer} on the development and test partition of the corpus.}
   \label{tab:reported_results_on_phoenix_1} \\
     \toprule
     &&&\multicolumn{4}{c|}{\stdsize Manuals}&
     \multicolumn{4}{c|}{\stdsize Non-M.}&
     \multicolumn{3}{c|}{\stdsize Fullframe}&
     \multicolumn{6}{c|}{Neural Architecture}&
     \multicolumn{2}{c|}{\acrshort{wer}}\\
     \centering \stdsize Reference & \stdsize Group& \stdsize Short Title&
   \rotatebox[origin=c]{90}{Location}&
   \rotatebox[origin=c]{90}{Movement}&
   \rotatebox[origin=c]{90}{Shape}&
   \rotatebox[origin=c]{90}{Orientation}&
   \rotatebox[origin=c]{90}{Head}&
   \rotatebox[origin=c]{90}{Mouth}&
   \rotatebox[origin=c]{90}{Eyes}&
   \rotatebox[origin=c]{90}{Eyebrows}&
   \rotatebox[origin=c]{90}{Bodyjoints}&
   \rotatebox[origin=c]{90}{RGB}&
   \rotatebox[origin=c]{90}{Motion}&
   \rotatebox[origin=c]{90}{CNN}&
   \rotatebox[origin=c]{90}{BLSTM}&
   \rotatebox[origin=c]{90}{LSTM}&
   \rotatebox[origin=c]{90}{BGRU}&
   \rotatebox[origin=c]{90}{Attention}&
   \rotatebox[origin=c]{90}{Transformer}&
  \multicolumn{1}{c}{Dev} &  \multicolumn{1}{c|}{Test} \\ \midrule\midrule
     \endfirsthead
 \multicolumn{22}{c}%
 {\stdsize\tablename\ \thetable{} -- continued from previous page} \\
     \toprule
     &&&\multicolumn{4}{c|}{\stdsize Manuals}&
     \multicolumn{4}{c|}{\stdsize Non-M.}&
     \multicolumn{3}{c|}{\stdsize Fullframe}&
     \multicolumn{6}{c|}{Neural Architecture}&
     \multicolumn{2}{c|}{\acrshort{wer}}\\
     \centering \stdsize Reference & \stdsize Group& \stdsize Short Title&
   \rotatebox[origin=c]{90}{Location}&
   \rotatebox[origin=c]{90}{Movement}&
   \rotatebox[origin=c]{90}{Shape}&
   \rotatebox[origin=c]{90}{Orientation}&
   \rotatebox[origin=c]{90}{Head}&
   \rotatebox[origin=c]{90}{Mouth}&
   \rotatebox[origin=c]{90}{Eyes}&
   \rotatebox[origin=c]{90}{Eyebrows}&
   \rotatebox[origin=c]{90}{Bodyjoints}&
   \rotatebox[origin=c]{90}{RGB}&
   \rotatebox[origin=c]{90}{Motion}&
   \rotatebox[origin=c]{90}{CNN}&
   \rotatebox[origin=c]{90}{BLSTM}&
   \rotatebox[origin=c]{90}{LSTM}&
   \rotatebox[origin=c]{90}{BGRU}&
   \rotatebox[origin=c]{90}{Attention}&
   \rotatebox[origin=c]{90}{Transformer}&
  \multicolumn{1}{c}{Dev} &  \multicolumn{1}{c|}{Test} \\ \midrule\midrule
 \endhead
 
 \midrule \multicolumn{22}{|r|}{{\stdsize Continued on next page}} \\ \bottomrule
 \endfoot
 
 \hline
 \endlastfoot

\cite{koller_continuous_2015} & RWTH & CSLR & x & x & x &  & x & x & x & x &  &  &  &  &  &  &  &  &  & 55.0&53.0\\
\cite{koller_automatic_2016} & RWTH/Surrey & Align Hamnosys & x & x & x & x & x & x & x & x &  &  &  & 2d &  &  &  &  &  & 49.6&48.2\\
\cite{koller_deep_2016} & RWTH/Surrey & 1 Million Hands & x & x & x &  & x & x & x & x &  &  &  & 2d &  &  &  &  &  & 47.1&45.1\\
\cite{koller16:hybridsign} & RWTH/Surrey & Deep Sign &  &  & x &  &  &  &  &  &  &  &  & 2d &  &  &  &  &  & 38.3&38.8\\
\cite{camgoz_subunets_2017} & Surrey/RWTH & SubUNets &  &  & x &  &  &  &  &  &  & x &  & 2d & x &  &  &  &  & 40.8&40.7\\
\cite{cui_recurrent_2017} & Tsinghua & Staged Optimization &  &  & x &  &  &  &  &  &  &  &  & 2d-1d & x &  &  &  &  & 39.4&38.7\\
\cite{koller_resign_2017} & RWTH & Re-Align &  &  &  &  &  &  &  &  &  & x &  & 2d & x &  &  &  &  & 27.1&26.8\\
\cite{huang_videobased_2018} & USTC/Here & Without Segmentation &  &  & x &  &  &  &  &  &  & x &  & 3d & x &  &  & x &  & -&38.3\\
\cite{wang_connectionist_2018} & Hefei Tech/USTC & Temporal Fusion &  &  &  &  &  &  &  &  &  & x &  & 3d-1d &  &  & x &  &  & 37.9&37.8\\
\cite{pu_dilated_2018} & USTC & Dilated Convolutions &  &  &  &  &  &  &  &  &  & x &  & 3d-Dilated &  &  &  &  &  & 38.0&37.3\\
\cite{koller_deep_2018} & RWTH/Surrey & Hybrid CNN-HMMs &  &  & x &  &  &  &  &  &  &  &  & 2d &  &  &  &  &  & 31.6&32.5\\
\cite{pei_continuous_2019} & Hefei Tech & Pseudo Supervised Learning &  &  &  &  &  &  &  &  &  & x &  & 3d &  &  & x &  &  & 40.9&40.6\\
\cite{song_parallel_2019} & Hefei Tech & Parallel Temp. Encoder &  &  &  &  &  &  &  &  &  & x &  & 3d-2d & x &  &  &  &  & 38.1&38.3\\
\cite{zhang_continuous_2019} & USTC & Reinforcement Learning &  &  &  &  &  &  &  &  &  & x &  & 3d &  &  &  &  & x & 38.0&38.3\\
\cite{cui_deep_2019} & Tsinghua & Iterative Training &  &  &  &  &  &  &  &  &  &  & x & 2d-1d & x &  &  &  &  & 37.9&37.6\\
\cite{pu_iterative_2019} & USTC & Iterative Alignment Network &  &  &  &  &  &  &  &  &  & x &  & 3d & x & x &  & x &  & 37.1&36.7\\
\cite{guo_dense_2019} & Hefei Tech/Huawei & Dense Temporal Conv. &  &  &  &  &  &  &  &  &  & x &  & 3d-1d &  &  &  &  &  & 35.9&36.5\\
\cite{yang_sfnet_2019} & Tencent/HKUST & SF-Net &  &  &  &  &  &  &  &  &  & x &  & 3d-2d & x &  &  &  &  & 35.6&34.9\\
\cite{zhou_dynamic_2019} & USTC & Pseudo Label Decoding &  &  &  &  &  &  &  &  &  & x &  & 3d-1d &  &  & x &  &  & 35.6&34.5\\
\cite{cui_deep_2019} & Tsinghua & Iterative Training &  &  & x &  &  &  &  &  &  &  &  & 2d-1d & x &  &  &  &  & 31.7&31.5\\
\cite{koller_weakly_2019} & RWTH/Surrey & Multi-Stream CNN-HMMs &  &  & x &  &  & x &  &  &  & x &  & 2d & x &  &  &  &  & 26.0&26.0\\
\cite{cui_deep_2019} & Tsinghua & Iterative Training &  &  &  &  &  &  &  &  &  & x &  & 2d-1d & x &  &  &  &  & 23.8&24.4\\
\cite{cui_deep_2019} & Tsinghua & Iterative Training &  &  &  &  &  &  &  &  &  & x & x & 2d-1d & x &  &  &  &  & 23.1&22.9\\
\cite{zhou_selfattentionbased_2020} & HKBU/HKU/BJTU/Nvidia & Fully-Inception Networks &  &  &  &  &  &  &  &  &  & x &  & 2d-1d &  &  &  &  & x & 31.7&31.3\\
\cite{adaloglou_comprehensive_2020} & CERTH/Patras & Comprehensive Study &  &  &  &  &  &  &  &  &  & x &  & 2d-1d & x &  &  &  &  & 28.9&29.1\\
\cite{borg_phonologicallymeaningful_2020} & Malta & Phonological Subunits &  & x & x &  &  &  &  &  & x &  &  &  & x &  &  &  &  & -&28.1\\
\cite{cheng_fully_2020} & HKUST/Tencent/Kwai & Fully Conv Networks &  &  &  &  &  &  &  &  &  & x &  & 2d-1d &  &  &  &  &  & 24.6&24.6\\
\cite{papastratis_continuous_2020} & CERTH & Cross-Modal Alignment &  &  &  &  &  &  &  &  &  & x &  & 2d-1d & x &  &  &  &  & 23.9&24.0\\
\cite{zhou_spatialtemporal_2020} & USTC & ST Multi-Cue Network &  &  & x &  & x &  &  &  & x & x &  & 2d-1d & x &  &  &  &  & 21.1&20.7\\

\end{longtable}
}
\end{center}

\renewcommand{\stdsize}{\footnotesize}
\newcolumntype{E}{>{\stdsize\collectcell\ColCell}p{0.5cm}<{\endcollectcell}} 
\newcolumntype{F}{>{\collectcell\ColCell\stdsize}m{1.5cm}<{\endcollectcell}}  
\newcolumntype{L}[1]{>{\RaggedRight\hspace{0pt}\scriptsize}p{#1}}
\newcolumntype{Q}[1]{>{\centering\stdsize}p{#1}}
\newcolumntype{P}[1]{>{\stdsize}p{#1}}
\newcolumntype{C}{>{\stdsize}{c}}
\newcolumntype{S}{>{\scriptsize}{c}}
\newcolumntype{R}{>{\stdsize}{r}}
\begin{center}
  \setlength{\tabcolsep}{1.5pt}
  { 
    \begin{longtable}{|L{2cm}|SS|C|CCCCCCCCCCC|CCCCCC|EE|}
     \caption{The table covers (to the best of our knowledge) all
       published sign language recognition works until mid 2020 that
       reported results on the RWTH-Phoenix Weather 2014
       \cite{koller_continuous_2015} task. The works are
       ordered by year and by \protect\gls{wer}. It allows to compare the
       type of employed data augmentation and the employed
       loss. Additionally, it can be seen if a paper performed an
       iterative training and the achieved performance in \gls{wer}}
   \label{tab:reported_results_on_phoenix_2} \\
   \toprule
   &&&
   &
     \multicolumn{11}{c|}{\stdsize Data Augmentation}&
     \multicolumn{6}{c|}{\stdsize Employed Loss}&
     \multicolumn{2}{c|}{\acrshort{wer}}\\
   \centering \stdsize Reference&\stdsize Group&\stdsize Short Title&
   \rotatebox[origin=c]{90}{Iterative Training}&
    \rotatebox[origin=c]{90}{Crop} &
    \rotatebox[origin=c]{90}{Framedrop} &
    \rotatebox[origin=c]{90}{Temporal Scaling} &
    \rotatebox[origin=c]{90}{Spatial Scaling} &
    \rotatebox[origin=c]{90}{Intenstity Noises} &
    \rotatebox[origin=c]{90}{Flip} &
    \rotatebox[origin=c]{90}{Brightness} &
    \rotatebox[origin=c]{90}{Contrast} &
    \rotatebox[origin=c]{90}{Hue} &
    \rotatebox[origin=c]{90}{Saturation} &
    \rotatebox[origin=c]{90}{Not Specified} &
    \rotatebox[origin=c]{90}{CE} &
    \rotatebox[origin=c]{90}{CTC} &
    \rotatebox[origin=c]{90}{KL-Divergence} &
    \rotatebox[origin=c]{90}{Squared Error} &
    \rotatebox[origin=c]{90}{Reinforce} &
    \rotatebox[origin=c]{90}{Other} &
    \multicolumn{1}{c}{Dev} & \multicolumn{1}{c|}{Test}\\ 
   \midrule\midrule
     \endfirsthead
 \multicolumn{23}{c}%
 {\stdsize\tablename\ \thetable{} -- continued from previous page} \\
 \toprule
    &&&
   &
     \multicolumn{11}{c|}{\stdsize Data Augmentation}&
     \multicolumn{6}{c|}{\stdsize Employed Loss}&
     \multicolumn{2}{c|}{\acrshort{wer}}\\
   \centering \stdsize Reference&\stdsize Group&\stdsize Short Title&
   \rotatebox[origin=c]{90}{Iterative Training}&
    \rotatebox[origin=c]{90}{Crop} &
    \rotatebox[origin=c]{90}{Framedrop} &
    \rotatebox[origin=c]{90}{Temporal Scaling} &
    \rotatebox[origin=c]{90}{Spatial Scaling} &
    \rotatebox[origin=c]{90}{Intenstity Noises} &
    \rotatebox[origin=c]{90}{Flip} &
    \rotatebox[origin=c]{90}{Brightness} &
    \rotatebox[origin=c]{90}{Contrast} &
    \rotatebox[origin=c]{90}{Hue} &
    \rotatebox[origin=c]{90}{Saturation} &
    \rotatebox[origin=c]{90}{Not Specified} &
    \rotatebox[origin=c]{90}{CE} &
    \rotatebox[origin=c]{90}{CTC} &
    \rotatebox[origin=c]{90}{KL-Divergence} &
    \rotatebox[origin=c]{90}{Squared Error} &
    \rotatebox[origin=c]{90}{Reinforce} &
    \rotatebox[origin=c]{90}{Other} &
    \multicolumn{1}{c}{Dev} & \multicolumn{1}{c|}{Test}\\ 
   \midrule\midrule
 \endhead
 
 \midrule \multicolumn{23}{|r|}{{\stdsize Continued on next page}} \\ \bottomrule
 \endfoot
 
 \hline
 \endlastfoot

\cite{koller_continuous_2015} & RWTH & CSLR &  &  &  &  &  &  &  &  &  &  &  &  &  &  &  &  &  &  & 55.0&53.0\\
\cite{koller_automatic_2016} & RWTH/Surrey & Align Hamnosys & x & x &  &  &  &  & x &  &  &  &  &  & x &  &  &  &  &  & 49.6&48.2\\
\cite{koller_deep_2016} & RWTH/Surrey & 1 Million Hands & x & x &  &  &  &  & x &  &  &  &  &  & x &  &  &  &  &  & 47.1&45.1\\
\cite{koller16:hybridsign} & RWTH/Surrey & Deep Sign &  & x &  &  &  &  & x &  &  &  &  &  & x &  &  &  &  &  & 38.3&38.8\\
\cite{camgoz_subunets_2017} & Surrey/RWTH & SubUNets & x &  &  &  &  &  &  &  &  &  &  & x &  & x &  &  &  &  & 40.8&40.7\\
\cite{cui_recurrent_2017} & Tsinghua & Staged Optimization & x &  &  & x &  &  &  &  &  &  &  &  &  & x & x &  &  & x & 39.4&38.7\\
\cite{koller_resign_2017} & RWTH & Re-Align & x & x &  &  &  &  & x &  &  &  &  &  & x &  &  &  &  &  & 27.1&26.8\\
\cite{huang_videobased_2018} & USTC/Here & Without Segmentation &  &  &  &  &  &  &  &  &  &  &  & x &  &  &  &  &  & x & -&38.3\\
\cite{wang_connectionist_2018} & Hefei Tech/USTC & Temporal Fusion &  &  &  &  &  &  &  &  &  &  &  & x &  & x &  &  &  &  & 37.9&37.8\\
\cite{pu_dilated_2018} & USTC & Dilated Convolutions & x &  &  &  &  &  &  &  &  &  &  & x & x & x &  &  &  &  & 38.0&37.3\\
\cite{koller_deep_2018} & RWTH/Surrey & Hybrid CNN-HMMs &  & x &  &  &  &  & x &  &  &  &  &  & x &  &  &  &  &  & 31.6&32.5\\
\cite{pei_continuous_2019} & Hefei Tech & Pseudo Supervised Learning & x &  &  &  &  &  &  &  &  &  &  & x &  & x &  &  &  &  & 40.9&40.6\\
\cite{song_parallel_2019} & Hefei Tech & Parallel Temp. Encoder & x &  &  &  &  &  &  &  &  &  &  & x &  & x &  & x &  &  & 38.1&38.3\\
\cite{zhang_continuous_2019} & USTC & Reinforcement Learning &  &  &  &  &  &  &  &  &  &  &  & x &  &  &  &  & x &  & 38.0&38.3\\
\cite{cui_deep_2019} & Tsinghua & Iterative Training & x &  &  & x & x & x &  &  &  &  &  &  &  & x & x &  &  & x & 37.9&37.6\\
\cite{pu_iterative_2019} & USTC & Iterative Alignment Network & x &  &  &  &  &  &  &  &  &  &  & x & x & x &  &  &  &  & 37.1&36.7\\
\cite{guo_dense_2019} & Hefei Tech/Huawei & Dense Temporal Conv. &  &  &  &  &  &  &  &  &  &  &  & x &  & x &  &  &  &  & 35.9&36.5\\
\cite{yang_sfnet_2019} & Tencent/HKUST & SF-Net &  & x &  &  &  &  &  &  &  &  &  &  &  & x & x &  &  &  & 35.6&34.9\\
\cite{zhou_dynamic_2019} & USTC & Pseudo Label Decoding & x &  &  &  &  &  &  &  &  &  &  & x & x & x & x &  &  &  & 35.6&34.5\\
\cite{cui_deep_2019} & Tsinghua & Iterative Training & x &  &  & x & x & x &  &  &  &  &  &  &  & x & x &  &  & x & 31.7&31.5\\
\cite{koller_weakly_2019} & RWTH/Surrey & Multi-Stream CNN-HMMs & x & x &  &  &  &  & x &  &  &  &  &  & x &  &  &  &  &  & 26.0&26.0\\
\cite{cui_deep_2019} & Tsinghua & Iterative Training & x &  &  & x & x & x &  &  &  &  &  &  &  & x & x &  &  & x & 23.8&24.4\\
\cite{cui_deep_2019} & Tsinghua & Iterative Training & x &  &  & x & x & x &  &  &  &  &  &  &  & x & x &  &  & x & 23.1&22.9\\
\cite{zhou_selfattentionbased_2020} & HKBU/HKU/BJTU/Nvidia & Fully-Inception Networks &  & x &  & x &  &  &  &  &  &  &  &  &  & x &  &  &  & x & 31.7&31.3\\
\cite{adaloglou_comprehensive_2020} & CERTH/Patras & Comprehensive Study &  & x & x &  &  &  &  & x & x & x & x &  & x & x &  &  &  & x & 28.9&29.1\\
\cite{borg_phonologicallymeaningful_2020} & Malta & Phonological Subunits & x &  &  &  &  &  &  &  &  &  &  &  & x & x &  &  &  &  & -&28.1\\
\cite{cheng_fully_2020} & HKUST/Tencent/Kwai & Fully Conv Networks &  & x &  & x &  &  &  &  &  &  &  &  & x & x &  &  &  &  & 24.6&24.6\\
\cite{papastratis_continuous_2020} & CERTH & Cross-Modal Alignment & x & x & x &  &  &  &  & x & x & x & x &  & x & x &  &  &  & x & 23.9&24.0\\
\cite{zhou_spatialtemporal_2020} & USTC & ST Multi-Cue Network & x & x & x &  &  &  & x &  &  &  &  &  & x & x &  &  &  & x & 21.1&20.7\\

\end{longtable}
}
\end{center}


\section{Conclusion and Outlook}
\label{sec:conclusion}

In this paper we shared, to the best of our knowledge, the most extensive quantitative study on the
field of sign language recognition covering analysis of over 300 publications from
1983 till 2020.
All analyzed studies have been manually tagged with a number of categories. This
source data is shared in the supplemental materials of this work.
Among others, we present following findings in this meta study:
\begin{itemize}
\item While many more studies are published on \gls{isolated} than on
  \gls{continuous} sign language recognition, the majority only covers small
  \gls{vocabulary} tasks.
\item After 2005 there was a paradigm shift in the community
  abandoning \gls{intrusive} capturing methods and embracing
  \gls{non-intrusive} methods.
\item Deep learning led the community towards the predominant use of global feature
  representations that are based on fullframe inputs. Those are
  particularly more common for larger \gls{vocabulary} tasks.
\item Non-manual parameters are still very rare in sign language
  recognition systems, despite their known importance for sign languages~\cite{pfau_nonmanuals_2010}. No sign recognition work has included eye gaze
  or blinks yet. Despite being the second most frequently
  researched sign language, research studies for \gls{csl} have hardly incorporated non-manual
  parameters. \Gls{dgs} is currently the only sign language where
  non-manuals have been successfully incorporated considering tasks
  with a \gls{vocabulary} of at least 200 signs.
  
\item \phoenix with a \gls{vocabulary} of 1080 signs 
represents the only resource for large vocabulary \gls{continuous} sign language world wide.
\end{itemize}

Moreover, we also presented the first meta analysis covering all known works that
compared themselves on the \phoenix benchmark data set.
Besides many details, we note that the best performing systems
typically adopt an iterative training style to overcome vanishing
gradients in deep \gls{cnn} architectures followed by
\glspl{blstm}. We also find that 2D convolutions followed by 1D
convolutions on fullframe inputs can be encountered in most state-of-the-art
systems. Surprisingly, we see that in many studies data augmentation
is not carefully described and also an ablation study that details the
effect of various augmentation methods is left for coming research.

We hope that in the future more works will include and be led by Deaf researchers, which
seems the only viable way to continue on this accelerated path the field is
currently on. More efforts are needed to create real-life large
vocabulary continuous sign language tasks that should be made
publicly accessible with well defined train, development and test partitions.

\printglossary[type=\acronymtype] 
\printglossary[type=main]
\bibliographystyle{apalike}
\bibliography{myLibrary}  



\clearpage
\section*{Supplemental Material}
\setcounter{table}{0}
\renewcommand{\thetable}{S\arabic{table}}

\newcolumntype{L}[1]{>{\RaggedRight\hspace{0pt}\stdsize}p{#1}}
\newcolumntype{Q}[1]{>{\centering\stdsize}p{#1}}
\newcolumntype{P}[1]{>{\stdsize}p{#1}}
\newcolumntype{C}{>{\stdsize}{c}}
\newcolumntype{R}{>{\stdsize}{r}}
\begin{center}
  \setlength{\tabcolsep}{1.8pt}
  { 
\begin{longtable}{|L{3.55cm}|C|RR|CC|CCCCC|CCCC|CCCCCC|CCCC|Q{1.5cm}|P{1.3cm}|}
      \caption{ The table covers most published works up until early 2020 in the
       field of sign language recognition  that
       report results on a sign language recognition task or introduce a
       related corpus.
       Each work is presented with a number of key
       properties. `Vocabulary' refers to the unique signs in a data
       set. `Signer' represents the joint number of signers in
       training and test sets. `Isolated' and
       `Continuous' refer to the nature of recorded sign
       language. `Input Data' specifies the employed capturing method
       for the input (`RGB' are colored images, `Depth' is depth
       information, `Colored' or `Electronic Gloves' represent
       \glspl{intrusive} gloves that ease the extraction of hand related
       features, `Mocap' means motion capture to accurately track the
       body parts). The three columns `Manuals', `Non-Manuals' and
       `Fullframe' characterize the employed features used for
       modeling. `Benchmark Dataset' specifies if the employed dataset
       is a known benchmark dataset, providing comparable tasks.
     }
   \label{tab:state_of_the_art_full_table} \\
     \toprule
     &&&&&&\multicolumn{5}{c|}{\stdsize Input Data}&\multicolumn{4}{c|}{\stdsize Manuals}&\multicolumn{6}{c|}{\stdsize Non-Manuals}&\multicolumn{4}{c|}{\stdsize Fullframe}&&\\
     \centering Reference &    Year &
   \rotatebox[origin=c]{90}{Vocabulary} &
   \rotatebox[origin=c]{90}{Signer} &
   \rotatebox[origin=c]{90}{Isolated} &
   \rotatebox[origin=c]{90}{Continuous} &
   \rotatebox[origin=c]{90}{RGB} &
   \rotatebox[origin=c]{90}{Depth} &
   \rotatebox[origin=c]{90}{Colored Glove} &
   \rotatebox[origin=c]{90}{Electronic Glove} &
   \rotatebox[origin=c]{90}{Mocap} &
   \rotatebox[origin=c]{90}{Location}&
   \rotatebox[origin=c]{90}{Movement}&
   \rotatebox[origin=c]{90}{Shape}&
   \rotatebox[origin=c]{90}{Orientation}&
   \rotatebox[origin=c]{90}{Head}&
   \rotatebox[origin=c]{90}{Mouth}&
   \rotatebox[origin=c]{90}{Eyes}&
   \rotatebox[origin=c]{90}{Eyeblink}&
   \rotatebox[origin=c]{90}{Eyebrows}&
   \rotatebox[origin=c]{90}{Eyegaze}&
   \rotatebox[origin=c]{90}{Bodyjoints}&
   \rotatebox[origin=c]{90}{RGB}&
   \rotatebox[origin=c]{90}{Depth}&
   \rotatebox[origin=c]{90}{Motion}&
   Benchmark Dataset&Language \\ \midrule\midrule
     \endfirsthead
 
 \multicolumn{12}{c}%
 {\tablename\ \thetable{} -- continued from previous page} \\
 \toprule
     &&&&&&\multicolumn{5}{c|}{\stdsize Input Data}&\multicolumn{4}{c|}{\stdsize Manuals}&\multicolumn{6}{c|}{\stdsize Non-Manuals}&\multicolumn{4}{c|}{\stdsize Fullframe}&&\\
     \centering Reference &    Year &
   \rotatebox[origin=c]{90}{Vocabulary} &
   \rotatebox[origin=c]{90}{Signer} &
   \rotatebox[origin=c]{90}{Isolated} &
   \rotatebox[origin=c]{90}{Continuous} &
   \rotatebox[origin=c]{90}{RGB} &
   \rotatebox[origin=c]{90}{Depth} &
   \rotatebox[origin=c]{90}{Colored Glove} &
   \rotatebox[origin=c]{90}{Electronic Glove} &
   \rotatebox[origin=c]{90}{Mocap} &
   \rotatebox[origin=c]{90}{Location}&
   \rotatebox[origin=c]{90}{Movement}&
   \rotatebox[origin=c]{90}{Shape}&
   \rotatebox[origin=c]{90}{Orientation}&
   \rotatebox[origin=c]{90}{Head}&
   \rotatebox[origin=c]{90}{Mouth}&
   \rotatebox[origin=c]{90}{Eyes}&
   \rotatebox[origin=c]{90}{Eyeblink}&
   \rotatebox[origin=c]{90}{Eyebrows}&
   \rotatebox[origin=c]{90}{Eyegaze}&
   \rotatebox[origin=c]{90}{Bodyjoints}&
   \rotatebox[origin=c]{90}{RGB}&
   \rotatebox[origin=c]{90}{Depth}&
   \rotatebox[origin=c]{90}{Motion}&
   Benchmark Dataset&Language \\ \midrule\midrule
 \endhead
 
 \midrule \multicolumn{27}{|r|}{{\stdsize Continued on next page}} \\ \bottomrule
 \endfoot
 
 \hline
 \endlastfoot

\cite{grimes_digital_1983} & 1983 & 26 &  & x &  &  &  &  & x &  &  &  & x & x &  &  &  &  &  &  &  &  &  &  &  & \acrshort{asl}\\
\cite{tamura_recognition_1988} & 1988 & 10 &  & x &  & x &  &  &  &  & x & x & x &  &  &  &  &  &  &  &  &  &  &  &  & \acrshort{jsl}\\
\cite{murakami_gesture_1991} & 1991 & 10 & 1 & x &  &  &  &  & x &  & x &  & x & x &  &  &  &  &  &  &  &  &  &  &  & \acrshort{jsl}\\
\cite{charayaphan_image_1992} & 1992 & 31 & 1 & x &  & x &  &  &  &  & x & x &  &  &  &  &  &  &  &  &  &  &  &  &  & \acrshort{asl}\\
\cite{fels_glovetalk_1993} & 1993 & 203 &  & x &  &  &  &  & x &  &  &  & x & x &  &  &  &  &  &  &  &  &  &  &  & \acrshort{asl}\\
\cite{kadous_grasp_1995} & 1995 & 95 & 5 & x &  &  &  &  & x &  & x & x & x & x &  &  &  &  &  &  &  &  &  &  &  & \acrshort{auslan}\\
\cite{liang_realtime_1995} & 1995 & 26 &  & x &  &  &  &  & x & x &  & x & x & x &  &  &  &  &  &  &  &  &  &  &  & \acrshort{asl}\\
\cite{waldron_isolated_1995} & 1995 & 14 &  & x &  &  &  &  & x & x & x & x & x & x &  &  &  &  &  &  &  &  &  &  &  & \acrshort{asl}\\
\end{longtable}
}
\end{center}


\end{document}